\documentclass[letterpaper,twocolumn,10pt]{article}
\usepackage{usenix2019_v3}
\usepackage{xspace}
\usepackage{setspace}
\usepackage{enumitem}

\usepackage{natbib}
\PassOptionsToPackage{numbers, compress, sort}{natbib}

\usepackage{tikz}
\usepackage{amsmath}

\usepackage{amsmath} 
\usepackage{amsfonts} 
\usepackage{amssymb}  
\usepackage[linesnumbered,ruled,vlined]{algorithm2e}
\usepackage{xcolor}
\usepackage{algcompatible}
\SetKwComment{Comment}{$\triangleright$ }{}
\SetKwInOut{Inputs}{Inputs}
\DontPrintSemicolon

\usepackage[table]{xcolor}
\usepackage{colortbl}
\usepackage{booktabs}
\usepackage{multirow}
\usepackage{adjustbox}
\usepackage{amsmath}
\usepackage{caption}
\usepackage{pgfplots}
\pgfplotsset{compat=newest}
\usepackage{xspace}
\usepackage{enumitem}
\usepackage{adjustbox}
\usepackage{scalefnt,sectsty}
\usepackage{tabularx}
\usepackage{ragged2e}

\usepackage{wrapfig}
\usepackage{cleveref}

\definecolor{HighlightGray}{gray}{0.85} 
\definecolor{StripeGray}{gray}{0.95}    
\usepackage{amsmath}
\usepackage{booktabs}     
\usepackage{siunitx}      
\usepackage[table]{xcolor} 
\usepackage{caption}      
\usepackage{natbib}       
\usepackage{multirow}     
\usepackage{makecell}     
\usepackage{rotating}     
\usepackage{siunitx}

\newcommand{\mypara}[1]{\vspace*{0.05in}\noindent\textbf{#1.} \xspace}

\newcommand{\ourmethod}{\textsc{WBC}\xspace}

\definecolor{boxbackground}{HTML}{FFFFFF} 
\definecolor{framecolor}{HTML}{D1D5DB}     
\definecolor{titletext}{HTML}{333333}      
\definecolor{titleline}{HTML}{AEB6BF}      

\usepackage{todonotes}

\begin{document}

\date{}

\title{\Large \bf Window-based Membership Inference Attacks Against \\Fine-tuned Large Language Models}


\author{
{\rm Yuetian Chen}\\
Purdue University
\and
{\rm Yuntao Du}\\
Purdue University
\and
{\rm Kaiyuan Zhang}\\
Purdue University
\and
{\rm Ashish Kundu}\\
Cisco Research
\and
{\rm Charles Fleming}\\
Cisco Systems
\and
{\rm Bruno Ribeiro}\\
Purdue University
\and
{\rm Ninghui Li}\\
Purdue University
} 

\maketitle


\begin{abstract}
Most membership inference attacks (MIAs) against Large Language Models (LLMs) rely on global signals, like average loss, to identify training data. This approach, however, dilutes the subtle, localized signals of memorization, reducing attack effectiveness. We challenge this global-averaging paradigm, positing that membership signals are more pronounced within localized contexts.
We introduce \ourmethod (Window-Based Comparison), which exploits this insight through a \emph{sliding window approach} with \emph{sign-based aggregation}. Our method slides windows of varying sizes across text sequences, with each window casting a binary vote on membership based on loss comparisons between target and reference models. By ensembling votes across geometrically spaced window sizes, we capture memorization patterns from token-level artifacts to phrase-level structures. Extensive experiments across eleven datasets demonstrate that \ourmethod substantially outperforms established baselines, achieving higher AUC scores and 2--3$\times$ improvements in detection rates at low false positive thresholds. Our findings reveal that aggregating localized evidence is fundamentally more effective than global averaging, exposing critical privacy vulnerabilities in fine-tuned LLMs.
\end{abstract}

\section{Introduction}
Large Language Models (LLMs) have achieved transformative success across numerous applications~\cite{grattafiori2024llama3herdmodels, qwen2025qwen25technicalreport, deepseekai2025deepseekv3technicalreport}. However, the training of these models often involves extensive datasets that can contain private or sensitive information. This practice introduces significant privacy risks, including the memorization and potential leakage of training data~\cite{carlini2021extracting, inan2021trainingdataleakageanalysis, huang2022largepretrainedlanguagemodels, li2024privacylargelanguagemodels}. Membership Inference Attacks (MIAs)—which determine whether specific samples were included in training data—are the primary method for quantifying these risks~\cite{shokri2017membership}, with successful attacks directly demonstrating information leakage~\cite{mireshghallah-etal-2022-empirical, fu2024, zeng2024exploring, puerto2025}.

\begin{figure}[t]
    \centering
    \includegraphics[width=0.9\linewidth]{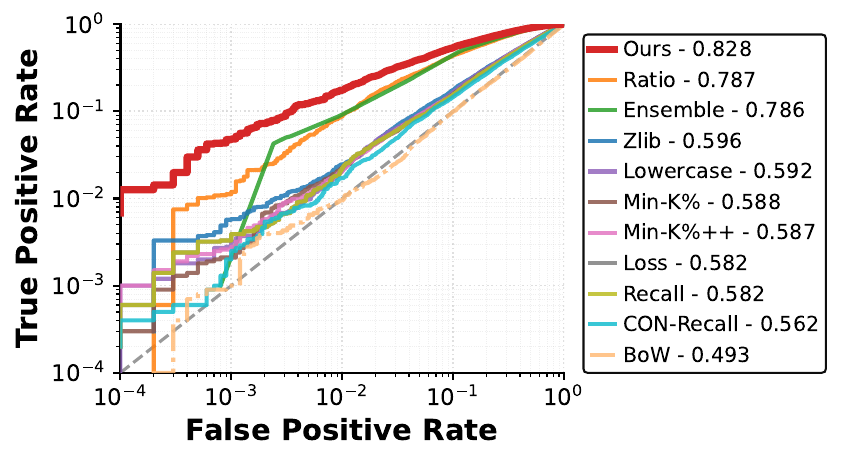}
    \caption{\textbf{Performance comparison of membership inference attacks on Pythia-2.8B.} ROC curves showing true positive rate vs. false positive rate (both on log scale) for various MIA methods evaluated on Web Samples V2 split. The diagonal dashed line represents random guessing performance. Numbers in legends indicate AUC scores. Our proposed \ourmethod Attack significantly outperforms existing baselines across all false positive rate regimes, demonstrating superior membership inference capability. }
    \label{fig:roc-overview}
\end{figure}

\begin{figure*}[t]
    \centering
    \includegraphics[width=0.9\linewidth]{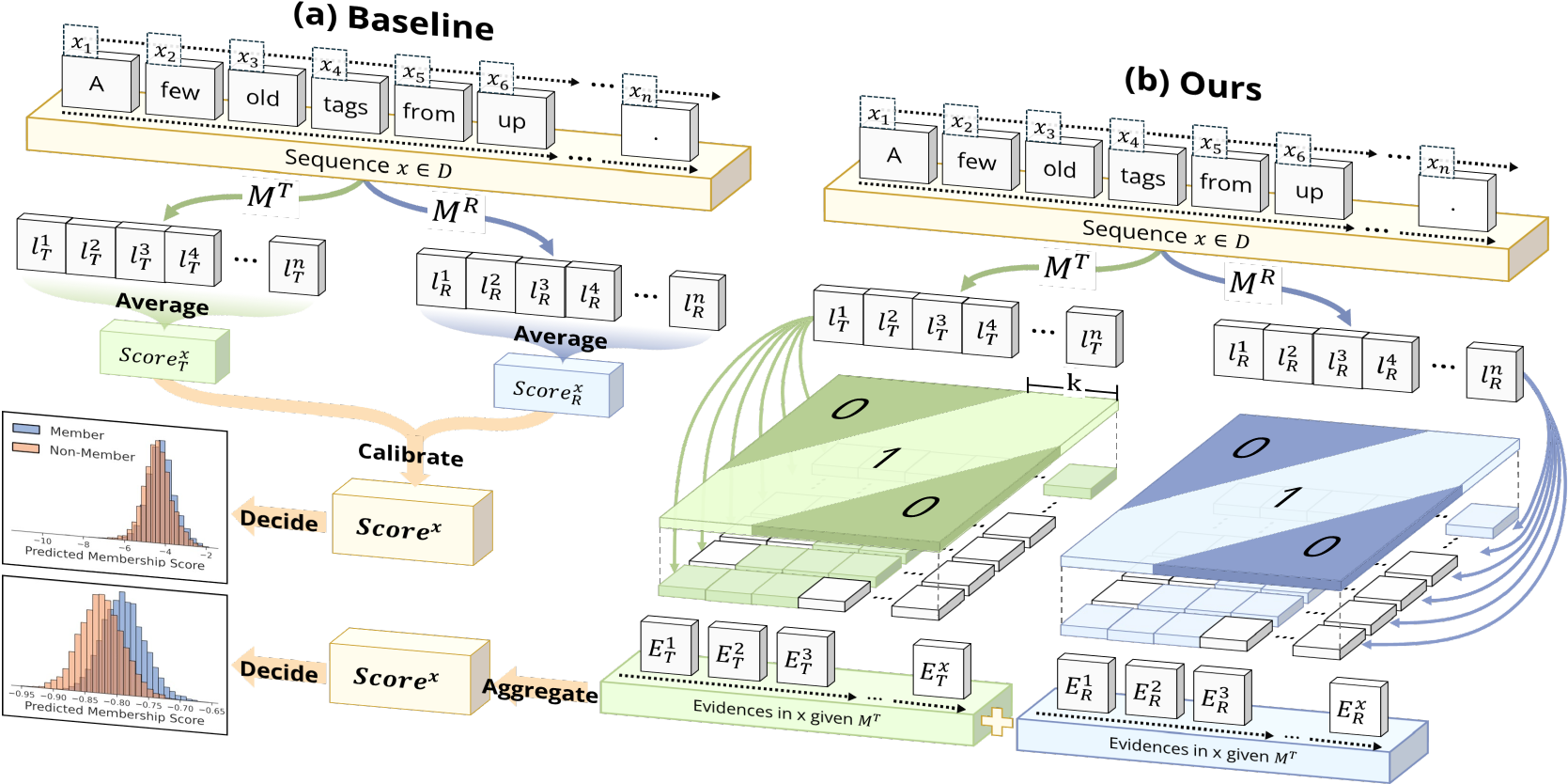}
    \caption{\textbf{Overview of the Window-Based Comparison (\ourmethod) Attack.} Unlike baseline methods \textbf{(a)} that rely on comparing a single, noisy global average of per-token losses, our approach \textbf{(b)} introduces a local aggregation step. We slide a window across the loss sequences from the target ($\mathcal{M}^\text{T}$) and reference ($\mathcal{M}^\text{R}$) models, making a binary comparison for each window. The final membership score is the sum of this local evidence, a process that filters noise and provides a more sensitive measure, leading to better separation between member and non-member distributions.}
    \label{fig:pipe}
\end{figure*}

These privacy vulnerabilities are particularly pronounced during the fine-tuning stage, where models are adapted to specialized, often proprietary, datasets~\cite{mireshghallah2022memorization,mireshghallah-etal-2022-empirical,chen2024janus}. Most established MIAs in this setting are reference-based, operating by comparing a global statistic---typically the average per-token loss---between a fine-tuned target model and a pre-trained reference model~\cite{watson2021importance, ye2022enhanced,fu2024,sok_ft_small_LM}.

In a reference-based MIA, for each instance
, one computes token-level losses (negative log predicted probability) from the reference model minus those from the target model.
To understand how to best utilize such token loss difference sequences for membership inference, we analyzed the distributions of loss differences for 10 million tokens. 
Our analysis reveals that membership signals manifest as sparse, extremal events rather than uniform distributional shifts.  Furthermore, these membership signals are intermixed with sparser and more extreme events caused by domain-specific tokens, which have high loss reduction and occur with similar frequencies in non-members as well as members.  These events create long-tailed noise with potentially infinite variance; global averaging becomes unreliable---a single outlier can dominate the entire statistic, obscuring genuine membership signals concentrated in localized regions.
This raises a critical question: 
\emph{Can robust detection methods overcome the fundamental challenge of extracting sparse membership signals from long-tailed noise dominated by domain adaptation effects?}

Building on the insights gained from the empirical analysis, we construct a mathematical model for the token loss reduction sequences using the point process theory from
extreme value statistics~\cite{leadbetter1983extremes,embrechts1997modelling}.  Theoretical analysis using the model shows that the most reliable membership signals are concentrated in localized token sequences. While individual token losses are too volatile to be trusted, aggregating them over short windows (e.g., 3-10 consecutive tokens) can filter noise without destroying the underlying signal. 

Taking advantage of these theoretical insights, we introduce the Window-Based Comparison (\ourmethod) attack, illustrated in Figure~\ref{fig:pipe}. Our method employs a \emph{sliding window approach} that performs hundreds of local comparisons, combined with \emph{sign-based aggregation} that counts the fraction of windows favoring membership. Our contributions are as follows:

\begin{enumerate}

    \item We are the first to use empirical analysis of distributions of token-level loss signals to understand how to construct more effective MIAs.  Our analysis results in several intriguing findings.  For example, counter-intuitively, the strongest membership signals occur on tokens where the fine-tuned model has a higher loss than the reference model.  See Section~\ref{subsubsec:structure_membership_sig} for discussion of these findings.  We conjecture that a similar analysis could result in useful insights for MIA in pre-trained LLMs and possibly other paradigms, including vision-language~\cite{radford2021learning} and diffusion language models~\cite{nie2025large}. 

    \item 
    We formalize these observations with 
    a mixture of point process models, which explains why global averaging is suboptimal and provides theoretical grounding for localized detection. We thus propose the \ourmethod attack that replaces global averaging with sliding window analysis. 
    Our method captures localized memorization patterns while maintaining robustness to long-tailed noise. A geometric ensemble strategy aggregates evidence across multiple window sizes, eliminating parameter tuning.
    
    \item Through extensive experiments on eleven diverse datasets using various models, we show that \ourmethod significantly outperforms thirteen baseline attacks. Averaged across all datasets, \ourmethod achieves an AUC of 0.839 compared to the strongest baseline's 0.754, and improves the True Positive Rate at 1\% False Positive Rate by $2.8\times$ (from 5.2\% to 14.6\%). Figure~\ref{fig:roc-overview} shows this superiority on a representative example, the Web Samples V2 dataset.
\end{enumerate}

Our findings prove that the aggregation of local signals is a more potent attack vector than previously established global methods. This work not only introduces a more effective MIA but also underscores the need for defenses that can account for these localized memorization patterns.
\section{Related Works}
\mypara{LLM} Large Language Models (LLMs) build on transformer architectures~\cite{vaswani2017attention}, employing self-supervised pre-training objectives such as masked language modeling~\cite{devlin2019bert}, causal language modeling~\cite{radford2019language}, or span corruption~\cite{raffel2020exploring} over vast, diverse corpora to capture linguistic patterns and world knowledge. Guided by empirical scaling laws that correlate model size, dataset volume, and compute with performance~\cite{kaplan2020scaling, hoffmann2022training, henighan2020scaling}, LLMs have scaled to trillions of parameters, exemplified by foundational models like GPT-3~\cite{brown2020language}, PaLM~\cite{chowdhery2023palm}, BLOOM~\cite{workshop2022bloom}, OPT~\cite{zhang2022opt}, LLaMA~\cite{touvron2023llama}, Mistral~\cite{jiang2023mistral7b}, and Falcon~\cite{penedo2023refinedweb}. Recent advancements incorporate architectural innovations, including mixture-of-experts (MoE) for efficient scaling~\cite{shazeer2017sparsely, fedus2022switch, jiang2024mixtral}, FlashAttention for optimized memory and speed~\cite{dao2205fast}, and hybrid designs blending dense and sparse activations~\cite{lepikhin2020gshard}. State-of-the-art models such as LLaMA-3~\cite{dubey2024llama}, Qwen3~\cite{yang2025qwen3}, DeepSeek-V3~\cite{deepseekai2025deepseekv3technicalreport}, and Gemma-3~\cite{team2025gemma} push boundaries in multilingual capabilities, reasoning, and efficiency. Despite their prowess in natural language understanding, generation, and downstream tasks~\cite{hendrycks2020measuring, liang2022holistic}, LLMs pose privacy risks through unintended memorization of sensitive training data, enabling extraction attacks and highlighting vulnerabilities in data curation~\cite{carlini2021extracting, inan2021trainingdataleakageanalysis, huang2022largepretrainedlanguagemodels, li2024privacylargelanguagemodels, nasr2023scalable}.

\mypara{MIA on LLMs} MIAs against LLMs initially showed limited success on pre-training data~\cite{duan2024, meeus2025sokmembershipinferenceattacks, das2025blind, satvaty2025}, attributed to massive datasets, few epochs, and fuzzy boundaries~\cite{carlini2021extracting, zhang2022text, karamolegkou2023copyright}. Early LLM MIAs adapted loss thresholding~\cite{yeom2018, shokri2017membership, sablayrolles2019white}, reference model calibration~\cite{watson2021importance, mireshghallah2022quantifying, lira}, and likelihood ratio tests~\cite{lira, ye2022enhanced,ndss26cpmia}. Fine-tuned LLMs demonstrated significantly higher vulnerability~\cite{mireshghallah-etal-2022-empirical, meeus2024did}, motivating specialized attacks: neighborhood comparison~\cite{mattern2023neighbor, Hisamoto_2020}, loss trajectory exploitation~\cite{li2022trajectory, li2024seqmia}, token probability analysis~\cite{shi2024, zhang2025}, and self-prompt calibration (SPV-MIA)~\cite{fu2024}. Advanced methods include instruction-based detection (MIA-Tuner)~\cite{fu2025mia}, user-level inference~\cite{kandpal2024, diwan2021fingerprintingfinetunedlanguagemodels}, context-aware attacks~\cite{wen2024membershipinferenceattacksincontext, chang2024context}, semantic-based approaches (SMIA)~\cite{mozaffari2024}, and alignment-specific attacks~\cite{feng2025exposingprivacygapsmembership}, and instruction-tuned models~\cite{hu2025membershipinferenceattacksvisionlanguage}. Label-only attacks~\cite{he2025labelonly, choquettechoo2021labelonly}, extraction methods~\cite{carlini2021extracting, nasr2023scalable}, and aggregation strategies~\cite{puerto2025} further expanded the threat landscape. 

\section{Preliminaries}
\label{sec:preliminaries}

This section provides the necessary background on autoregressive language models, establishing the formulation for per-token loss that underpins our attack. It then formally defines the threat model, detailing the adversary's objective, knowledge, and capabilities.

\subsection{Autoregressive Language Model}

\mypara{Next-Token Prediction}
The dominant paradigm for Large Language Models (LLMs) is autoregressive modeling \citep{radford2019language, brown2020language}. For a sequence of discrete tokens $\mathbf{x} = (x_1, x_2, \ldots, x_n)$, these models operate on the unidirectional dependency assumption, where the probability of observing a token $x_i$ depends only on its prefix $(x_1, \ldots, x_{i-1})$. This permits factorization of the joint probability as:
\begin{equation}
\label{eq:autoregressive_factorization}
p(\mathbf{x}) = \prod_{i=1}^{n} p(x_i \mid x_1, \ldots, x_{i-1})
\end{equation}
A model $\mathcal{M}$ with parameters $\theta$ is trained to maximize this likelihood over a large corpus, typically by minimizing the cross-entropy loss. The core metric derived from this process, and the fundamental signal for our attack, is the per-token loss (negative log-likelihood) for a given sequence:
\begin{equation}
\label{eq:per_token_loss}
\ell_i^{\mathcal{M}} = -\log p_{\mathcal{M}}(x_i \mid x_1, \ldots, x_{i-1})
\end{equation}
This value, $\ell_i^{\mathcal{M}}$, quantifies the model's ``surprise'' at seeing token $x_i$ given the preceding context.

\mypara{Fine-Tuning and Memorization}
While pre-trained on vast, general-domain corpora, LLMs are often specialized for downstream tasks via fine-tuning on smaller, targeted datasets. This process adapts the model's parameters to the new data distribution~\cite{radford2018improving,devlin2019bert}. A well-documented side effect of fine-tuning is that it amplifies the model's memorization of the training samples~\cite{mireshghallah2022quantifying, zeng2024exploring}. Consequently, for a data sample $\mathbf{x}$ that is part of the fine-tuning set (a ``member''), the target model $\mathcal{M}^{\text{T}}$ will exhibit significantly lower average loss compared to its loss on unseen data. Reference-based attacks leverage this phenomenon by comparing the target model's loss against that of a reference model $\mathcal{M}^{\text{R}}$ (typically the pre-trained base model), which has not been exposed to the fine-tuning data. The discrepancy in loss reduction between members and non-members provides a potent signal for membership inference.

\subsection{Threat Model and Attacker Capabilities}

\mypara{Adversary's Objective}
We investigate Membership Inference Attacks (MIAs) in the context of fine-tuned LLMs. The adversary's goal is to determine if a specific data record $\mathbf{x}$ was part of the private fine-tuning dataset, $D_{\text{train}}$. Given a target model $\mathcal{M}^\text{T}$ fine-tuned on $D_{\text{train}}$, the adversary constructs an attack function $A(\mathbf{x}, \mathcal{M}^\text{T}, \mathcal{M}^\text{R})$ that outputs a real-valued score or a binary prediction $\hat{m}(\mathbf{x}) \in \{0, 1\}$, where $\hat{m}(\mathbf{x}) = 1$ signifies a prediction that $\mathbf{x} \in D_{\text{train}}$.

\mypara{Access Model and Knowledge}
We assume a \textit{score-based black-box} access model, a realistic scenario for attacks against deployed LLMs where internal model details are inaccessible~\citep{shi2024,zhang2024pretraining,zhang2025}. In this setting, the adversary's capabilities are limited to querying the target model $\mathcal{M}^{\text{T}}$ with a text sequence $\mathbf{x}$ and receiving only the corresponding sequence of per-token loss values $\{\ell_i^{\text{T}}\}_{i=1}^n$. This distinguishes our setting from stricter black-box scenarios where only sequence-level averaged losses are returned~\citep{shokri2017membership,watson2021importance,mireshghallah2022quantifying,fu2024,Huang_Liu_He_Li_2025,meng2025rr}. This assumption reflects two standard deployment scenarios: (1) \textit{Open-weight adaptation}, where practitioners fine-tune public models (e.g., via HuggingFace~\cite{wolf2019huggingface}) on proprietary data, granting attackers access to the model weights for local inference; and (2) \textit{API-based inference}, where standard serving backends like vLLM~\citep{kwon2023efficient} explicitly support parameters such as \texttt{prompt\_logprobs}, returning the exact signal required by \ourmethod. Furthermore, consistent with reference-based attack literature~\citep{watson2021importance,fu2024}, the adversary is assumed to have identical score-based black-box access to a reference model, $\mathcal{M}^\text{R}$. The most principled choice for this reference is the pre-trained base model from which $\mathcal{M}^\text{T}$ was fine-tuned, as this best isolates the memorization signal induced by the fine-tuning process. As shown in \Cref{app:misaligned_references}, \ourmethod remains robust and continues to outperform baselines even when $\mathcal{M}^\text{R}$ is a misaligned model. The adversary has no access to the model's internal components, such as its parameters, gradients, or hidden-state activations, nor to the membership status of any sample. 
\section{Window-Based Comparison Attack}

A fundamental limitation of prior membership inference attacks is their reliance on global statistics that aggregate losses over entire texts. This document-level averaging is dominated by extremal events, rare tokens with outlier losses that overwhelm genuine membership signals. Our empirical analysis reveals that these outliers can be 10--100 times larger than typical fluctuations, making global statistics unreliable for detecting the sparse, localized memorization patterns that distinguish members from non-members. This motivates our window-based approach: by evaluating hundreds of local comparisons instead of a single global average, we can isolate membership signals from contaminating noise.

\ourmethod attack exploits this clustering by systematically evaluating contiguous text windows. Rather than computing a single global statistic, we perform hundreds of local comparisons and aggregate their outcomes. This approach transforms the noisy, high-dimensional problem of token-level analysis into a robust voting mechanism across multiple granularities.

\subsection{Theoretical Foundation: Extremal Events and Window-Based Detection}
\label{subsec:theory}

In this section, we provide a plausible explanation for why our sign-based localized window-based detection outperforms global averaging for membership inference. Our analysis of over 10 million token-level comparisons reveals that membership signals are governed by a regime of extremal events. These extremes arise from two sources: (1) tokens where the fine-tuned model achieves dramatic perplexity reduction compared to the reference model, appearing in both members and non-members due to domain adaptation, and (2) membership-specific tokens where fine-tuning provides additional confidence due to memorization. The former creates extreme values in the loss difference distribution that dominate global averages but carry no membership information, masking the discriminative signals from the latter. We argue that this behavior aligns with a mixture of point processes, where non-informative extremal events from domain adaptation obscure the sparse true membership signals. Our sliding window approach isolates these localized patterns, and our sign-based binary aggregation strategy provides robustness to the long-tailed distributions created by both types of extremes.


\subsubsection{Empirical Structure of Membership Signals}
\label{subsubsec:structure_membership_sig}

To understand the nature of membership signals in fine-tuned LLMs, we conducted an empirical analysis of token-level loss differences. We fine-tuned Pythia-2.8B~\cite{biderman2023pythia} on the Khan Academy subset of Cosmopedia~\cite{benallal2024cosmopedia}, following standard practices with a learning rate of $5 \times 10^{-5}$ and training for 3 epochs on 10,000 samples. We then evaluated both the fine-tuned target model $\mathcal{M}^{\text{T}}$ and the original pre-trained reference model $\mathcal{M}^{\text{R}}$ on balanced sets of 10,000 member samples (from the training set) and 10,000 non-member samples (held-out data from the same distribution).

For each text sequence $\mathbf{x} = (x_1, \ldots, x_n)$, we computed the per-token loss (negative log-likelihood) for both models, yielding loss sequences $\{\ell_j^{\text{T}}\}_{j=1}^n$ and $\{\ell_j^{\text{R}}\}_{j=1}^n$. The token-level loss difference $\Delta_j = \ell_j^{\text{R}} - \ell_j^{\text{T}}$ quantifies how much more confident the fine-tuned model is compared to the reference model at each position. Positive values indicate positions where fine-tuning improved prediction.

\begin{figure}[t]
    \centering
    \includegraphics[width=\linewidth]{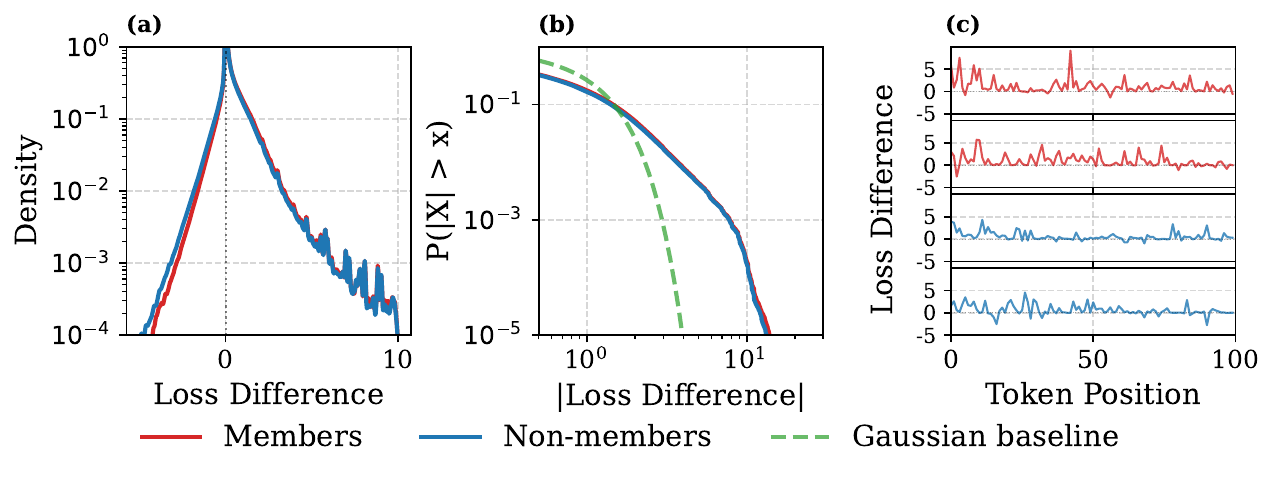}
    \caption{\textbf{Empirical distribution of token-level loss differences.} (a) Log-scale density plots reveal long-tailed distributions with a subtle rightward shift for members. The difference is most pronounced in the left tail, not the mean. (b) Complementary CDF confirms the long-tailed behavior of the token loss. (c) Time series shows sparse, scattered extremes.}
    \label{fig:loss_diff_dist}
\end{figure}

\vspace*{0.05in}\noindent\textbf{Figure~\ref{fig:loss_diff_dist}(a)} presents the empirical distribution of these loss differences on a logarithmic scale to emphasize tail behavior. The distributions reveal several critical insights. 

First, the curves for members and non-members look quite similar, but there is a small but non-trivial mean difference between members (0.393) and non-members (0.331). 
This suggests that using global averaging for membership inference would work to some limited extent.  In order to achieve higher attack effectiveness, we need to identify and extract finer-grained membership signals from the token-loss sequences. Furthermore, the overwhelming majority of tokens show near-zero differences, suggesting that any fine-grained membership signals would appear as relatively rare events.

Second, the right tail regions for both members and non-members extend to very high positive values and are almost overlapping.  This suggests that \emph{tokens with dramatic loss reduction are not good membership signals}.  While this observation may appear counterintuitive, it makes perfect sense.  For tokens to result in high loss reduction, they must appear many times during fine-tuning, meaning that they are domain-specific features that appear frequently across the dataset, in non-members as well as members.  We call these \emph{domain-specific tokens}.  High loss reduction on these tokens are the result of domain adaptation that occurs because of fine-tuning.  

This fact creates significant challenges for distinguishing members from non-members.  For example, the approach of using the maximum loss reduction across all tokens to determine membership would not be effective.

Third, the most noticeable difference between the two distributions is that the one for members shows a small but consistent rightward shift in the left tail region. 
As this region consists of tokens on which the fine-tuned model actually performs worse than the reference model, it counter-intuitively suggests that \emph{the strongest membership signals appear in tokens where the target model has higher loss}.  

We believe that the reasons for this phenomenon are as follows. First, for a token to be a good signal of membership of an instance, it should appear in this instance and few other instances.  As a result, we expect this token's loss to be just slightly lower compared to the case that the instance is a non-member.  Second, as the fine-tuned model needs to increase predicted probabilities for domain-specific features, it would have to reduce predicted probabilities for some other tokens.  Since members are used in fine-tuning, such reductions, when they occur, would be less than those for non-members.  

Compared to the right-tail region, the left-tail region extends less horizontally, but has higher density.  This suggests that such membership signals have less magnitude than the noise signals from the right region, but are less rare. 
 
Lastly, both member and non-member distributions exhibit long tails with excess kurtosis exceeding 18, far surpassing the value of 0 expected for Gaussian distributions. They also show positive skewness (2.82 for members, 2.63 for non-members), confirming the asymmetric nature of these extremal events. This means that the extremal events from both left-tail and right-tail regions will occur frequently enough to be exploited for membership inference. 
These observations motivate modeling membership signals as a mixture of extremal event processes rather than a uniform or Gaussian process. 


\vspace*{0.05in}\noindent\textbf{Figure~\ref{fig:loss_diff_dist}(b)} shows the log-log scale of the token loss complementary cumulative distribution (i.e., 1-CDF).  It reveals a long-tailed behavior, a hallmark of distributions whose averages are dominated by rare extreme events. For members, approximately 1.77\% of tokens exceed three standard deviations from the mean, compared to only 0.3\% expected under a Gaussian distribution. 

\vspace*{0.05in}\noindent\textbf{Figure~\ref{fig:loss_diff_dist}(c)} illustrates the sequential structure of these signals by plotting loss differences across token positions for representative samples. Instead of clustered patterns suggestive of memorization of contiguous passages, we observe sparse, scattered spikes distributed seemingly at random throughout the sequences. Statistical analysis confirms this observation: computing the clustering coefficient (ratio of observed to expected spacing between extreme values)~\cite{clark1954distance,ripley1976second} yields values of 1.049 for members and 1.053 for non-members, where 1.0 indicates perfectly random placement under a Poisson point process~\cite{diggle2013statistical}. This absence of spatial clustering suggests that membership signals manifest as isolated, extremal events rather than coherent, memorized passages. 


\subsubsection{Modeling Membership Signals as Extremal Events}

The empirical observations we made above from Figure~\ref{fig:loss_diff_dist} collectively suggest that membership inference should be framed not as detecting distribution shifts through averages but as identifying and aggregating evidence from sparse extremal events---a perspective grounded in point process theory from extreme value statistics~\cite{leadbetter1983extremes,embrechts1997modelling}.
From these observations, we model the per-token loss difference point process as a superposition of three components:
\begin{equation}
\Delta_j(\mathbf{x}) = \ell_j^\text{R} - \ell_j^\text{T} = \mathbb{I}[\mathbf{x} \in D_{\text{train}}] \cdot \delta_j(\mathbf{x}) + \xi_j + \epsilon_j,
\end{equation}
Here $\epsilon_j$ represents baseline noise with a small mean $\mu_\epsilon$ and variance $\sigma^2$, capturing typical prediction fluctuations. 
The term $\xi_j$ captures \emph{domain-specific tokens} (e.g., frequent technical terms) that exhibit high loss reduction due to domain adaptation independent of membership. Since these features appear in both members and non-members, $\xi_j$ acts as high-magnitude noise that dominates the right tail of the distribution. We model these events due to domain-specific tokens as a point process where events occur with low probability but large magnitude. Specifically, $\xi_j = Z_j \cdot |Y_j|$ where $Z_j \sim \text{Bernoulli}(\rho_\xi)$ with $\rho_\xi \ll 1$ marks rare token occurrence, and $Y_j$ follows a long-tailed distribution representing the magnitude of these events. From our observations, approximately 1.8\% of tokens exhibit extreme values exceeding $3\sigma$, suggesting $\rho_\xi \approx 0.02$. 
The term $\delta_j(\mathbf{x}) \geq 0$ represents the membership signal at position $j$. Unlike domain features, these signals manifest as smaller, more frequent loss reductions resulting from the memorization of specific instances, causing  $\Delta_j(\mathbf{x})$ to increase when $\mathbf{x}$ is a member as opposed to a non-member. 

Similarly, membership signals ($\delta_j(\mathbf{x})$'s) follow a sparse pattern: $\delta_j(\mathbf{x}) = B_j \cdot \gamma_j$ where $B_j \sim \text{Bernoulli}(\rho_\delta)$ indicates whether position $j$ contains memorized content, and $\gamma_j > 0$ represents the signal strength when present.



\mypara{The Masking Effect of Extremal Events in Global Averages} Under our model, global averaging reveals why traditional membership inference approaches struggle with fine-tuned LLMs. The global average loss difference across all tokens becomes:

\begin{equation}
\bar{\Delta} = \frac{1}{n}\sum_{j=1}^n \Delta_j = \underbrace{\mathbb{I}[\mathbf{x} \in D_{\text{train}}] \cdot \rho_\delta \bar{\gamma}}_{\text{membership signal}} +\!\!\! \underbrace{\frac{1}{n}\sum_{j=1}^n \xi_j}_{\text{rare token noise}} \!\!\!+ \!\!\!\!\!\!\underbrace{\bar{\epsilon}}_{\text{baseline noise}}\!\!\!\!\!\!\!.
\end{equation}
The first term represents the true membership signal with expected value $\rho_\delta \bar{\gamma}$. However, the second term, arising from rare token events, has variance proportional to $\rho_\xi \cdot \mathbb{E}[Y_j^2]$. For long-tailed distributions of $Y_j$, this variance can be extremely large. Hence, a single extreme $\xi_j$ value can dominate the entire average, as these outliers can be 10-100 times larger than typical fluctuations.

This simple model explains why global averaging methods can have weak MIA signals: the signal-to-noise ratio deteriorates not due to the weakness of membership signals, but because rare token events create overwhelming noise that cannot be averaged away. With only $\rho_\xi n \approx 0.02n$ extreme events in a sequence, the law of large numbers converges slowly (or not at all for some long-tailed distributions), making global statistics unreliable signals.

\subsubsection{Window-Based Detection and Robust Aggregation}

Instead of attempting to detect a global mean shift corrupted by long-tailed noise, we reformulate membership inference as detecting localized extremal events. This formulation aligns with standard problems in local sequence analysis, where the design space ranges from change-point detection algorithms to various aggregation functions. However, unlike change-point detection, which typically seeks contiguous regime shifts, our empirical analysis at \Cref{subsubsec:structure_membership_sig} indicates that membership signals do not appear as such sharp transitions; rather, they manifest as continuous, low-degree signals that are consistently buried under the dominated signal of domain adaptation. We therefore adopt a sliding window approach as a particularly effective instance of this paradigm to capture these scattered events. For a window of size $w$ starting at position $i$, we compute the windowed sum:
\begin{equation}
S_i(w) = \sum_{j=i}^{i+w-1} \Delta_j,
\end{equation}
where the summation spans $w$ consecutive tokens from position $i$ to position $i+w-1$. By sliding this window across the entire sequence, we obtain $n-w+1$ different windows.

The distribution of these windowed sums follows a contaminated mixture:
\begin{equation}\label{eq:si}
S_i(w) \sim (1 - p_{\text{extreme}}) \cdot \mathcal{F}_{\text{normal}} + p_{\text{extreme}} \cdot \mathcal{F}_{\text{long}}
\end{equation}
where $p_{\text{extreme}}$ is the probability that a window contains either a membership signal or a rare token event, and $\mathcal{F}_{\text{long}}$ represents a long-tailed distribution. Consider two natural approaches to aggregate these windowed differences into a membership score:
\begin{align}
T_{\text{mean}} &= \frac{1}{n-w+1} \sum_{i=1}^{n-w+1} S_i(w), & \text{(mean-based)}\\
T_{\text{sign}} &= \frac{1}{n-w+1} \sum_{i=1}^{n-w+1} \mathbb{I}[S_i(w) > 0]. & \text{(sign-based)}
\end{align}
The mean-based approach uses the magnitude of each window's difference for MIA, while the sign-based approach uses only the direction. Classical robust statistics theory shows which method offers superior statistical power and under which conditions. The Pitman asymptotic relative efficiency (ARE) quantifies the relative sample sizes needed by two tests to achieve equal power~\cite{pitman1949notes,lehmann2005testing}. For location testing under contaminated distributions, the ARE is given by
\begin{equation}
\text{ARE}(\text{sign}, \text{mean}) = 4f^2(0) \cdot \text{Var}[S_i(w)],
\end{equation}
where $f(0)$ is the density of the centered distribution at zero and $\text{Var}[S_i(w)]$ is the variance of $S_i(w)$. When ARE exceeds unity, the sign test requires fewer samples than the mean test for equal power~\cite{van2000asymptotic}.

For the long-tailed contaminated mixture in \Cref{eq:si}, $\text{Var}[S_i(w)]$ is dominated by the long-tailed extremal events $\mathcal{F}_{\text{long}}$. Meanwhile, the density $f(0)$ remains bounded because most windows cluster near zero. Consequently, the ARE grows with contamination degree, strongly favoring the sign test. 
To illustrate this phenomenon, let's examine an extreme scenario: an unrealistic long-tailed distribution, such as Cauchy contamination. In this case, the ARE can be infinite, implying that the mean test has zero asymptotic efficiency relative to the sign test~\cite{huber2011robust}. This result underscores the robustness of the sign test when variance is dominated by extremes.
This theoretical result directly motivates our use of the sign test:
\begin{align}
T_{\text{sign}} &= \frac{1}{n-w+1} \sum_{i=1}^{n-w+1} \mathbb{I}[S_i(w) > 0]\\
&= \frac{1}{n-w+1} \sum_{i=1}^{n-w+1} \mathbb{I}\left[\sum_{j=i}^{i+w-1} \ell_j^{\text{R}} > \sum_{j=i}^{i+w-1} \ell_j^{\text{T}}\right].
\end{align}
This statistic counts the fraction of windows where the reference model loss exceeds the target model loss, regardless of difference magnitude. By adapting this standard non-parametric test to the domain of membership inference, we inherit several key known robustness guarantees:

\mypara{Breakdown point} The breakdown point of an estimator is the largest fraction of data that can be arbitrarily corrupted without the estimator becoming uninformative~\cite{donoho1983notion,huber2011robust}. The sign test achieves the maximum possible breakdown point of 0.5, meaning it remains reliable even when up to half the windows contain arbitrarily large contaminating values from rare tokens. In our context, ``window contamination'' refers to windows where rare token events $\xi_j$ create extreme values that dominate the window sum. Even if 50\% of windows contain such extreme contaminations, the sign test still correctly identifies the majority vote, whereas the mean would be completely dominated by these outliers.

\mypara{Scale invariance} The sign test is invariant to monotone transformations of the data~\cite{lehmann2005testing}. Whether the loss differences are measured in nats, bits, or any monotone transformation thereof, the sign test yields identical results. Unlike the mean, which is sensitive to non-linear scaling, this invariance ensures consistent detection regardless of calibration differences.

\mypara{Bounded output} Unlike the mean, which can take arbitrary values depending on outlier magnitudes, the sign statistic is naturally bounded in [0,1], representing the fraction of windows favoring membership. This bounded range facilitates consistent threshold selection across different datasets and models without requiring dataset-specific normalization.

The fundamental distinction is that mean-based aggregation attempts to measure ``how much lower is the loss on average,'' while sign-based comparison asks ``how often is the loss lower'', a more robust question under long-tailed noise. For empirically observed contamination levels ($p_{\text{extreme}} \approx 0.05-0.10$), the ARE typically exceeds 2 to 5, meaning sign-based comparison requires 2 to 5 times fewer samples than mean aggregation for equivalent detection power. 
This explains our empirical results in Section~\ref{subsubsec:aggregation}, where sign-based aggregation consistently outperforms mean, median, and min aggregation across all datasets, with particularly pronounced advantages in high-precision regimes.
\subsubsection{Window Size Trade-off and Ensemble Strategy}

The use of windows in detecting membership signals introduces a trade-off, governed by the window size $w$. On one hand, smaller windows (e.g., $w=1$ or $2$) increase the number of (correlated) tests that can be performed, as there are $n-w+1$ possible windows in a sequence of length $n$. However, this comes at the cost of a poor signal-to-noise ratio for each individual window sum $S_i(w) = \sum_{j=i}^{i+w-1} \Delta_j$. Specifically, the expected signal scales with $\rho_\delta w \bar{\gamma}$, while the standard deviation due to baseline noise scales with $\sqrt{w}\sigma$. For very small $w$, the probability that a window containing a signal yields $S_i(w) > 0$ is barely above 0.5, making each binary test uninformative despite the large number of tests available.

Conversely, excessively large window sizes introduce three key challenges. In probability theory~\cite{Naus:1982:ADS}, our task is related to the well-known {\em scan statistic tests}, and the effect of $w$ is well-documented. First, the {\em effective sample size} of any statistics diminishes sharply with window size $w$. 
Although the total number of windows is $n - w + 1$, which decreases only linearly with $w$, adjacent windows share $w - 1$ tokens, inducing strong dependencies. Consequently, the effective number of independent tests is closer to $n / w$, the maximum number of non-overlapping windows, which decays rapidly as $w$ increases.

Second, large windows elevate the {\em risk of contamination by rare token events}. When a window contains an extreme rare token event $\xi_j$, its magnitude can dominate the sum $S_i(w)$, overshadowing the accumulated membership signal. The probability of such contamination is proportional to $1 - (1 - \rho_\xi)^w \approx w\rho_\xi$ for small $\rho_\xi$, where $\rho_\xi$ is the rare token probability.

Third, {\em signal dilution} becomes pronounced. In a window of size $w$, only $\rho_\delta w$ tokens are expected to be memorized, while the remaining $(1 - \rho_\delta)w$ tokens are non-memorized. For sparse signals ($\rho_\delta \ll 1$), this averaging dilutes the membership signal, reducing the likelihood that windows containing signals yield positive sums.
Empirically, an intermediate window (3–10 tokens) balances these competing factors. However, determining the optimal window size analytically requires parameters that cannot be estimated reliably: the signal sparsity $\rho_\delta$, rare token frequency $\rho_\xi$, signal strength $\bar{\gamma}$, and rare token distribution $\mathbb{E}[Y^2]$. These parameters vary across datasets and even within documents---e.g., technical sections exhibit different patterns than narrative prose. We provide a detailed analysis of this optimization problem in Appendix~\ref{app:window_optimization}.

Thus, instead of pursuing an elusive optimal window size, we adopt an ensemble approach that provides robustness through diversification. We employ a geometric progression that densely samples small windows while maintaining coverage of larger scales:
\begin{equation}
\label{eq:window_progression}
w_k = \text{round} \left( w_{\min} \cdot \left( \frac{w_{\max}}{w_{\min}} \right)^{\frac{k-1}{|W|-1}} \right)
\end{equation}
where $k \in \{1, \ldots, |W|\}$, and $w_{\min}$, $w_{\max}$ bound the range of scales. The geometric spacing ensures that the ratio between consecutive window sizes remains approximately constant: $w_{k+1}/w_k \approx (w_{\max}/w_{\min})^{1/(|W|-1)}$. This provides equal relative resolution across scales, a principle from scale-space theory that optimizes multi-resolution analysis. Small windows, where memorization signals are most likely, receive dense sampling, while larger windows are sampled more sparsely to provide coverage without redundancy. The final ensemble score uses uniform weights, yielding
\begin{equation}
S_{\text{\ourmethod}} = \frac{1}{|W|} \sum_{k=1}^{|W|} T_{\text{sign}}(w_k).
\end{equation}
Uniform weighting follows the principle of maximum entropy—minimizing worst-case regret under parameter uncertainty. This equal weighting also provides variance reduction through averaging partially correlated measurements, with ensemble variance decreasing as $\tau^2[1 + (|W| - 1)\rho]/|W|$ when correlations $\rho$ are modest.

This equal weighting is theoretically justified under parameter uncertainty. By the principle of maximum entropy, uniform weights minimize worst-case regret when the true optimal window size is unknown. Additionally, uniform averaging provides variance reduction: if individual window scores have variance $\tau^2$ and correlation $\rho$, the ensemble variance becomes $\tau^2[1 + (|W| - 1)\rho]/|W|$, which decreases with ensemble size when correlations are modest.

The ensemble strategy solves two fundamental problems. First, it provides robustness against unknown parameters ($\rho_\delta$, $\rho_\xi$, $\bar{\gamma}$, $\mathbb{E}[Y^2]$) that vary across datasets and even within documents. Second, it captures heterogeneous memorization patterns that no single window size can detect optimally—small windows (2-4 tokens) capture token-level artifacts, medium windows (5-10 tokens) detect phrases, and larger windows identify paragraph-level patterns. When texts mix technical terms with quoted passages, the ensemble naturally combines these complementary detection capabilities.

This ensemble consistently outperforms the empirically best single window size, as validated in our ablation studies (Sections~\ref{subsubsec:ensemble}). The strategy thus achieves both robustness and superior performance—transforming theoretical insights into a practical attack requiring no parameter tuning.

\subsection{Algorithmic Specification}

\begin{algorithm}[t]
\caption{\ourmethod Attack}
\label{alg:wbc}

\Inputs{Target model $\mathcal{M}^\text{T}$; Reference model $\mathcal{M}^\text{R}$; Input $\mathbf{x}=\{x_j\}_{j=1}^n$; Window size schemes $W$}

\ForAll{$k \in \{\text{T}, \text{R}\}$}{
  \For{$j=1$ \KwTo $n$}{
    $\ell^k_j \gets -\log p_{\mathcal{M}^k}(x_j \mid x_1, \dots, x_{j-1})$
  }
}

$S_{\text{\ourmethod}} \gets 0$

\ForAll{$w \in W$}{
  $sum^{\text{T}} \gets 0$; $sum^{\text{R}} \gets 0$
  \For{$j=1$ \KwTo $w$}{
    $sum^{\text{T}} \gets sum^{\text{T}} + \ell^{\text{T}}_j$
    $sum^{\text{R}} \gets sum^{\text{R}} + \ell^{\text{R}}_j$
  }
  \Comment{Initialize sums for first window}
  $\texttt{count} \gets 0$
  \If{$sum^{\text{R}} > sum^{\text{T}}$}{
    $\texttt{count} \gets \texttt{count} + 1$
  }
  \For{$i=2$ \KwTo $n-w+1$}{
    $sum^{\text{T}} \gets sum^{\text{T}} - \ell^{\text{T}}_{i-1} + \ell^{\text{T}}_{i+w-1}$
    
    $sum^{\text{R}} \gets sum^{\text{R}} - \ell^{\text{R}}_{i-1} + \ell^{\text{R}}_{i+w-1}$
    
    \If{$sum^{\text{R}} > sum^{\text{T}}$}{
      $\texttt{count} \gets \texttt{count} + 1$
    }
  }
  \Comment{Slide \& compare}
  $T_{\text{sign}}(w) \gets \texttt{count} / (n - w + 1)$
  
  $S_{\text{\ourmethod}} \gets S_{\text{\ourmethod}} + T_{\text{sign}}(w)$
}

$S_{\text{\ourmethod}} \gets S_{\text{\ourmethod}} / |W|$

\Return $S_{\text{\ourmethod}}$
\end{algorithm}

The complete procedure for our attack is detailed in Algorithm \ref{alg:wbc}. The procedure unfolds as follows. First, the per-token negative log-likelihoods (i.e., the negative log-probability of each token conditioned on its prefix) are computed for the input $\mathbf{x}$ using both the target ($\mathcal{M}^\text{T}$) and reference ($\mathcal{M}^\text{R}$) models, yielding the loss sequences $\{\ell_j^{\text{T}}\}_{j=1}^n$ and $\{\ell_j^{\text{R}}\}_{j=1}^n$ (Lines 1-3). The algorithm then iterates through each window size $w \in W$ (Line 5). For a given $w$, it initializes the sums for the first window (Lines 6-7) and performs the first sign-based comparison (Lines 8-9). It then slides the window from the second position to the end, incrementally updating the sums and the comparison count at each step (Lines 10-14). Once all windows for size $w$ are processed, the normalized count yields the window-specific score $T_{\text{sign}}(w)$ (Line 15), which is added to the total ensemble score (Line 16). Finally, this total is averaged over the number of window sizes $|W|$ to produce the final score $S_{\text{\ourmethod}}$ (Lines 17-18). In practice, we further accelerate this computation by leveraging optimized convolution implementations. Details and performance evaluation are provided in Appendix~\ref{app:convolution}.

\section{Experiments}
\begin{table*}[t]
    \centering
    \footnotesize
    \scalefont{0.95}
    \setlength{\tabcolsep}{2.5pt} 
    \rowcolors{6}{white}{StripeGray}

    \caption{MIA performance (AUC, TPR@10\%FPR, TPR@1\%FPR, TPR@0.1\%FPR) across different datasets. Each cell shows mean with std. dev. as a subscript. Best-performing results are highlighted. We observe that \ourmethod consistently outperforms all baseline methods across all reported metrics and datasets.}
    \label{tab:main_results_part_one}
    \begin{tabular}{@{} l *{3}{r r r r} @{}}
        \toprule
        \multirow{2}{*}{\textbf{MIAs}} &
        \multicolumn{4}{c}{\textbf{Khan Academy}} &
        \multicolumn{4}{c}{\textbf{Stanford}} &
        \multicolumn{4}{c}{\textbf{Stories}} \\

        \cmidrule(lr){2-5} \cmidrule(lr){6-9} \cmidrule(lr){10-13}

        & {AUC} & {T@10\%} & {T@1\%} & {T@0.1\%} & {AUC} & {T@10\%} & {T@1\%} & {T@0.1\%} & {AUC} & {T@10\%} & {T@1\%} & {T@0.1\%} \\
        \midrule

        Loss~\citep{yeom2018}           & 0.568\textsubscript{$\pm$.003} & 0.151\textsubscript{$\pm$.003} & 0.017\textsubscript{$\pm$.002} & 0.001\textsubscript{$\pm$.001} & 0.590\textsubscript{$\pm$.006} & 0.170\textsubscript{$\pm$.008} & 0.023\textsubscript{$\pm$.003} & 0.003\textsubscript{$\pm$.001} & 0.587\textsubscript{$\pm$.004} & 0.146\textsubscript{$\pm$.008} & 0.020\textsubscript{$\pm$.001} & 0.003\textsubscript{$\pm$.001} \\
        ZLIB~\citep{carlini2021extracting}   & 0.583\textsubscript{$\pm$.004} & 0.161\textsubscript{$\pm$.003} & 0.019\textsubscript{$\pm$.003} & 0.003\textsubscript{$\pm$.001} & 0.606\textsubscript{$\pm$.006} & 0.178\textsubscript{$\pm$.009} & 0.024\textsubscript{$\pm$.003} & 0.002\textsubscript{$\pm$.000} & 0.596\textsubscript{$\pm$.004} & 0.157\textsubscript{$\pm$.005} & 0.019\textsubscript{$\pm$.002} & 0.003\textsubscript{$\pm$.001} \\
        Lowercase~\citep{carlini2021extracting} & 0.586\textsubscript{$\pm$.004} & 0.154\textsubscript{$\pm$.006} & 0.013\textsubscript{$\pm$.002} & 0.002\textsubscript{$\pm$.001} & 0.596\textsubscript{$\pm$.003} & 0.173\textsubscript{$\pm$.004} & 0.024\textsubscript{$\pm$.002} & 0.003\textsubscript{$\pm$.001} & 0.593\textsubscript{$\pm$.004} & 0.172\textsubscript{$\pm$.004} & 0.018\textsubscript{$\pm$.002} & 0.002\textsubscript{$\pm$.001} \\
        Min-K\%~\citep{shi2024}           & 0.595\textsubscript{$\pm$.002} & 0.182\textsubscript{$\pm$.004} & 0.024\textsubscript{$\pm$.002} & 0.002\textsubscript{$\pm$.000} & 0.593\textsubscript{$\pm$.005} & 0.168\textsubscript{$\pm$.006} & 0.024\textsubscript{$\pm$.002} & 0.003\textsubscript{$\pm$.001} & 0.588\textsubscript{$\pm$.003} & 0.145\textsubscript{$\pm$.004} & 0.019\textsubscript{$\pm$.003} & 0.003\textsubscript{$\pm$.001} \\
        Min-K\%++~\citep{zhang2025}       & 0.596\textsubscript{$\pm$.002} & 0.180\textsubscript{$\pm$.005} & 0.022\textsubscript{$\pm$.003} & 0.002\textsubscript{$\pm$.001} & 0.592\textsubscript{$\pm$.004} & 0.161\textsubscript{$\pm$.006} & 0.023\textsubscript{$\pm$.002} & 0.004\textsubscript{$\pm$.001} & 0.586\textsubscript{$\pm$.003} & 0.150\textsubscript{$\pm$.006} & 0.018\textsubscript{$\pm$.002} & 0.002\textsubscript{$\pm$.001} \\
        BoWs~\cite{das2025blind}           & 0.499\textsubscript{$\pm$.004} & 0.101\textsubscript{$\pm$.003} & 0.011\textsubscript{$\pm$.001} & 0.001\textsubscript{$\pm$.001} & 0.498\textsubscript{$\pm$.003} & 0.109\textsubscript{$\pm$.003} & 0.012\textsubscript{$\pm$.001} & 0.001\textsubscript{$\pm$.001} & 0.502\textsubscript{$\pm$.005} & 0.100\textsubscript{$\pm$.002} & 0.011\textsubscript{$\pm$.001} & 0.002\textsubscript{$\pm$.001} \\
        ReCall~\citep{xie2024recall}       & 0.568\textsubscript{$\pm$.003} & 0.151\textsubscript{$\pm$.003} & 0.017\textsubscript{$\pm$.002} & 0.001\textsubscript{$\pm$.001} & 0.590\textsubscript{$\pm$.006} & 0.170\textsubscript{$\pm$.008} & 0.023\textsubscript{$\pm$.003} & 0.003\textsubscript{$\pm$.001} & 0.587\textsubscript{$\pm$.004} & 0.146\textsubscript{$\pm$.008} & 0.020\textsubscript{$\pm$.001} & 0.003\textsubscript{$\pm$.001} \\
        CON-Recall~\citep{wang2025conrecalldetectingpretrainingdata} & 0.563\textsubscript{$\pm$.003} & 0.142\textsubscript{$\pm$.004} & 0.014\textsubscript{$\pm$.002} & 0.002\textsubscript{$\pm$.001} & 0.570\textsubscript{$\pm$.006} & 0.159\textsubscript{$\pm$.007} & 0.018\textsubscript{$\pm$.002} & 0.003\textsubscript{$\pm$.000} & 0.558\textsubscript{$\pm$.004} & 0.143\textsubscript{$\pm$.006} & 0.020\textsubscript{$\pm$.002} & 0.002\textsubscript{$\pm$.001} \\
        DC-PDD~\cite{zhang2024pretraining} & 0.567\textsubscript{$\pm$.003} & 0.136\textsubscript{$\pm$.003} & 0.015\textsubscript{$\pm$.002} &
        0.002\textsubscript{$\pm$.001} & 0.570\textsubscript{$\pm$.003} & 0.140\textsubscript{$\pm$.003} & 0.014\textsubscript{$\pm$.002} & 0.002\textsubscript{$\pm$.001} & 0.574\textsubscript{$\pm$.002} & 0.146\textsubscript{$\pm$.002} & 0.019\textsubscript{$\pm$.002} & 0.002\textsubscript{$\pm$.001} \\
        SPV-MIA~\cite{fu2024} & 0.695\textsubscript{$\pm$.003} & 0.240\textsubscript{$\pm$.006} & 0.049\textsubscript{$\pm$.004} & 0.005\textsubscript{$\pm$.003} & 0.760\textsubscript{$\pm$.003} & 0.260\textsubscript{$\pm$.010} & 0.077\textsubscript{$\pm$.005} & 0.012\textsubscript{$\pm$.003} & 0.763\textsubscript{$\pm$.004} & 0.360\textsubscript{$\pm$.007} & 0.070\textsubscript{$\pm$.006} & 0.018\textsubscript{$\pm$.005} \\
        Ratio~\cite{watson2021importance} & 0.703\textsubscript{$\pm$.003} & 0.264\textsubscript{$\pm$.004} & 0.037\textsubscript{$\pm$.004} & 0.003\textsubscript{$\pm$.001} & 0.781\textsubscript{$\pm$.004} & 0.401\textsubscript{$\pm$.010} & 0.080\textsubscript{$\pm$.008} & 0.013\textsubscript{$\pm$.001} & 0.769\textsubscript{$\pm$.004} & 0.389\textsubscript{$\pm$.011} & 0.091\textsubscript{$\pm$.008} & 0.022\textsubscript{$\pm$.006} \\
        Difference~\cite{watson2021importance} & 0.692\textsubscript{$\pm$.002} & 0.259\textsubscript{$\pm$.005} & 0.045\textsubscript{$\pm$.002} & 0.003\textsubscript{$\pm$.001} & 0.742\textsubscript{$\pm$.005} & 0.360\textsubscript{$\pm$.011} & 0.080\textsubscript{$\pm$.008} & 0.021\textsubscript{$\pm$.004} & 0.719\textsubscript{$\pm$.005} & 0.348\textsubscript{$\pm$.007} & 0.079\textsubscript{$\pm$.006} & 0.014\textsubscript{$\pm$.003} \\
        Ensemble~\cite{zhang2025softselectivedataobfuscation}        & 0.687\textsubscript{$\pm$.003} & 0.217\textsubscript{$\pm$.004} & 0.061\textsubscript{$\pm$.002} & 0.007\textsubscript{$\pm$.001} & 0.738\textsubscript{$\pm$.004} & 0.142\textsubscript{$\pm$.003} & 0.075\textsubscript{$\pm$.006} & 0.011\textsubscript{$\pm$.004} & 0.758\textsubscript{$\pm$.004} & 0.338\textsubscript{$\pm$.004} & 0.048\textsubscript{$\pm$.003} & 0.014\textsubscript{$\pm$.003} \\
        \midrule
        \rowcolor{HighlightGray} 
        \textbf{\ourmethod} (Ours)           & \bfseries 0.837\textsubscript{$\pm$.003} & \bfseries 0.538\textsubscript{$\pm$.007} & \bfseries 0.146\textsubscript{$\pm$.009} & \bfseries 0.026\textsubscript{$\pm$.009} & \bfseries 0.854\textsubscript{$\pm$.003} & \bfseries 0.583\textsubscript{$\pm$.008} & \bfseries 0.194\textsubscript{$\pm$.012} & \bfseries 0.034\textsubscript{$\pm$.017} & \bfseries 0.808\textsubscript{$\pm$.005} & \bfseries 0.494\textsubscript{$\pm$.007} & \bfseries 0.160\textsubscript{$\pm$.013} & \bfseries 0.034\textsubscript{$\pm$.003} \\
        \bottomrule
        \rowcolor{white}
        & & & & & & & & & & & & \\
    
        \toprule
        \multirow{2}{*}{\textbf{MIAs}} &
        \multicolumn{4}{c}{\textbf{Web Samples v2}} &
        \multicolumn{4}{c}{\textbf{Auto Math Text}} &
        \multicolumn{4}{c}{\textbf{WikiHow}} \\

        \cmidrule(lr){2-5} \cmidrule(lr){6-9} \cmidrule(lr){10-13}

        & {AUC} & {T@10\%} & {T@1\%} & {T@0.1\%} & {AUC} & {T@10\%} & {T@1\%} & {T@0.1\%} & {AUC} & {T@10\%} & {T@1\%} & {T@0.1\%} \\
        \midrule

        Loss~\citep{yeom2018}           & 0.579\textsubscript{$\pm$.004} & 0.153\textsubscript{$\pm$.006} & 0.022\textsubscript{$\pm$.003} & 0.004\textsubscript{$\pm$.001} & 0.553\textsubscript{$\pm$.005} & 0.133\textsubscript{$\pm$.005} & 0.014\textsubscript{$\pm$.001} & 0.002\textsubscript{$\pm$.001} & 0.592\textsubscript{$\pm$.004} & 0.162\textsubscript{$\pm$.005} & 0.022\textsubscript{$\pm$.002} & 0.004\textsubscript{$\pm$.001} \\
        ZLIB~\citep{carlini2021extracting}   & 0.593\textsubscript{$\pm$.004} & 0.175\textsubscript{$\pm$.005} & 0.023\textsubscript{$\pm$.002} & 0.005\textsubscript{$\pm$.001} & 0.562\textsubscript{$\pm$.005} & 0.141\textsubscript{$\pm$.005} & 0.015\textsubscript{$\pm$.002} & 0.002\textsubscript{$\pm$.001} & 0.604\textsubscript{$\pm$.004} & 0.171\textsubscript{$\pm$.004} & 0.022\textsubscript{$\pm$.002} & 0.002\textsubscript{$\pm$.000} \\
        Lowercase~\citep{carlini2021extracting} & 0.592\textsubscript{$\pm$.004} & 0.166\textsubscript{$\pm$.005} & 0.021\textsubscript{$\pm$.003} & 0.003\textsubscript{$\pm$.001} & 0.568\textsubscript{$\pm$.003} & 0.141\textsubscript{$\pm$.008} & 0.015\textsubscript{$\pm$.002} & 0.001\textsubscript{$\pm$.001} & 0.614\textsubscript{$\pm$.004} & 0.182\textsubscript{$\pm$.007} & 0.024\textsubscript{$\pm$.002} & 0.001\textsubscript{$\pm$.001} \\
        Min-K\%~\citep{shi2024}            & 0.587\textsubscript{$\pm$.005} & 0.158\textsubscript{$\pm$.007} & 0.023\textsubscript{$\pm$.002} & 0.002\textsubscript{$\pm$.000} & 0.581\textsubscript{$\pm$.006} & 0.150\textsubscript{$\pm$.006} & 0.021\textsubscript{$\pm$.001} & 0.002\textsubscript{$\pm$.000} & 0.600\textsubscript{$\pm$.004} & 0.178\textsubscript{$\pm$.005} & 0.025\textsubscript{$\pm$.002} & 0.004\textsubscript{$\pm$.001} \\
        Min-K\%++~\citep{zhang2025}        & 0.586\textsubscript{$\pm$.004} & 0.153\textsubscript{$\pm$.005} & 0.020\textsubscript{$\pm$.002} & 0.003\textsubscript{$\pm$.001} & 0.583\textsubscript{$\pm$.005} & 0.150\textsubscript{$\pm$.005} & 0.022\textsubscript{$\pm$.002} & 0.003\textsubscript{$\pm$.001} & 0.598\textsubscript{$\pm$.004} & 0.174\textsubscript{$\pm$.005} & 0.027\textsubscript{$\pm$.003} & 0.004\textsubscript{$\pm$.001} \\
        BoWs~\cite{das2025blind}            & 0.493\textsubscript{$\pm$.004} & 0.097\textsubscript{$\pm$.004} & 0.010\textsubscript{$\pm$.001} & 0.002\textsubscript{$\pm$.001} & 0.500\textsubscript{$\pm$.004} & 0.105\textsubscript{$\pm$.005} & 0.009\textsubscript{$\pm$.001} & 0.001\textsubscript{$\pm$.000} & 0.498\textsubscript{$\pm$.004} & 0.096\textsubscript{$\pm$.003} & 0.008\textsubscript{$\pm$.001} & 0.001\textsubscript{$\pm$.000} \\
        ReCall~\citep{xie2024recall}        & 0.579\textsubscript{$\pm$.004} & 0.153\textsubscript{$\pm$.006} & 0.022\textsubscript{$\pm$.003} & 0.004\textsubscript{$\pm$.001} & 0.553\textsubscript{$\pm$.005} & 0.133\textsubscript{$\pm$.005} & 0.014\textsubscript{$\pm$.001} & 0.002\textsubscript{$\pm$.001} & 0.592\textsubscript{$\pm$.004} & 0.162\textsubscript{$\pm$.005} & 0.022\textsubscript{$\pm$.002} & 0.004\textsubscript{$\pm$.001} \\
        CON-Recall~\citep{wang2025conrecalldetectingpretrainingdata} & 0.560\textsubscript{$\pm$.004} & 0.144\textsubscript{$\pm$.003} & 0.017\textsubscript{$\pm$.001} & 0.002\textsubscript{$\pm$.001} & 0.547\textsubscript{$\pm$.005} & 0.135\textsubscript{$\pm$.005} & 0.015\textsubscript{$\pm$.001} & 0.002\textsubscript{$\pm$.001} & 0.571\textsubscript{$\pm$.005} & 0.161\textsubscript{$\pm$.005} & 0.019\textsubscript{$\pm$.002} & 0.002\textsubscript{$\pm$.001} \\
        DC-PDD~\cite{zhang2024pretraining} & 0.569\textsubscript{$\pm$.003} & 0.143\textsubscript{$\pm$.003} & 0.019\textsubscript{$\pm$.002} &
        0.003\textsubscript{$\pm$.001} & 0.560\textsubscript{$\pm$.006} & 0.130\textsubscript{$\pm$.006} & 0.015\textsubscript{$\pm$.002} & 0.002\textsubscript{$\pm$.001} & 0.571\textsubscript{$\pm$.004} & 0.143\textsubscript{$\pm$.003} & 0.016\textsubscript{$\pm$.002} & 0.002\textsubscript{$\pm$.001} \\
        SPV-MIA~\cite{fu2024} & 0.787\textsubscript{$\pm$.003} & 0.456\textsubscript{$\pm$.006} & 0.088\textsubscript{$\pm$.004} & 0.008\textsubscript{$\pm$.006} & 0.759\textsubscript{$\pm$.003} & 0.369\textsubscript{$\pm$.006} & 0.063\textsubscript{$\pm$.006} & 0.007\textsubscript{$\pm$.004} & 0.719\textsubscript{$\pm$.003} & 0.300\textsubscript{$\pm$.007} & 0.053\textsubscript{$\pm$.005} & 0.013\textsubscript{$\pm$.004} \\
        Ratio~\cite{watson2021importance} & 0.788\textsubscript{$\pm$.003} & 0.435\textsubscript{$\pm$.009} & 0.090\textsubscript{$\pm$.007} & 0.012\textsubscript{$\pm$.003} & 0.768\textsubscript{$\pm$.002} & 0.378\textsubscript{$\pm$.005} & 0.075\textsubscript{$\pm$.002} & 0.011\textsubscript{$\pm$.006} & 0.714\textsubscript{$\pm$.003} & 0.261\textsubscript{$\pm$.006} & 0.042\textsubscript{$\pm$.003} & 0.008\textsubscript{$\pm$.002} \\
        Difference~\cite{watson2021importance} & 0.739\textsubscript{$\pm$.003} & 0.365\textsubscript{$\pm$.008} & 0.094\textsubscript{$\pm$.008} & 0.019\textsubscript{$\pm$.007} & 0.700\textsubscript{$\pm$.004} & 0.300\textsubscript{$\pm$.009} & 0.051\textsubscript{$\pm$.006} & 0.009\textsubscript{$\pm$.003} & 0.709\textsubscript{$\pm$.003} & 0.278\textsubscript{$\pm$.005} & 0.047\textsubscript{$\pm$.003} & 0.009\textsubscript{$\pm$.002} \\
        Ensemble~\cite{zhang2025softselectivedataobfuscation}                            & 0.786\textsubscript{$\pm$.003} & 0.477\textsubscript{$\pm$.004} & 0.086\textsubscript{$\pm$.002} & 0.005\textsubscript{$\pm$.013} & 0.749\textsubscript{$\pm$.002} & 0.359\textsubscript{$\pm$.004} & 0.051\textsubscript{$\pm$.002} & 0.003\textsubscript{$\pm$.002} & 0.724\textsubscript{$\pm$.003} & 0.338\textsubscript{$\pm$.004} & 0.065\textsubscript{$\pm$.005} & 0.018\textsubscript{$\pm$.001} \\
        \midrule
        \rowcolor{HighlightGray} 
        \textbf{\ourmethod} (Ours)               & \bfseries 0.843\textsubscript{$\pm$.003} & \bfseries 0.573\textsubscript{$\pm$.008} & \bfseries 0.198\textsubscript{$\pm$.012} & \bfseries 0.045\textsubscript{$\pm$.004} & \bfseries 0.814\textsubscript{$\pm$.003} & \bfseries 0.504\textsubscript{$\pm$.007} & \bfseries 0.153\textsubscript{$\pm$.009} & \bfseries 0.040\textsubscript{$\pm$.008} & \bfseries 0.802\textsubscript{$\pm$.003} & \bfseries 0.451\textsubscript{$\pm$.010} & \bfseries 0.096\textsubscript{$\pm$.007} & \bfseries 0.019\textsubscript{$\pm$.006} \\
        \bottomrule
    \end{tabular}
\end{table*}

This section details the empirical evaluation of our proposed \ourmethod attack. We first describe the experimental setup, including the datasets, models, and baseline attacks. We then present our main results, demonstrating the superior performance of our method, followed by ablation studies and analyses that provide deeper insights into why our approach is effective.

\subsection{Experimental Setup}

\mypara{Datasets} We evaluate on eleven datasets spanning synthetic and real-world domains. The first category consists of six subsets from Cosmopedia~\cite{benallal2024cosmopedia}, a large-scale synthetic dataset generated by Mixtral-8x7B-Instruct-v0.1~\cite{jiang2024mixtral}. We utilize the Khan Academy, Stanford, Stories, Web Samples v2, AutoMathText, and WikiHow subsets as these represent the scale and quality of data commonly used in modern fine-tuning practices, where practitioners increasingly rely on high-quality synthetic data for domain adaptation. The second category comprises real-world document benchmarks, including WikiText-103~\cite{merity2016pointer} and XSum~\cite{narayan2018don} for direct comparison with prior work~\cite{fu2024}, supplemented by Amazon Reviews~\cite{ni2019justifying}, CC-News~\cite{Hamborg2017}, and Reddit~\cite{reddit_title_body_dataset} to extend evaluation across diverse domains at similar scales. All evaluations use balanced 10,000-sample splits for members/non-members with a minimum 512-token length. See Appendix~\ref{app:datasets} for detailed descriptions.


\mypara{Models and Fine-Tuning} Our primary analysis is conducted on the Pythia-2.8B~\cite{biderman2023pythia}. To assess the generalizability of our findings, we perform additional experiments across a wide range of model architectures and scales. To study the effect of model scale, we evaluate on the Pythia Scaling Suite~\cite{biderman2023pythia}, including the 160M, 410M, 1B, 1.4B, and 6.9B parameter models. To test architectural diversity, we also evaluate on several widely adopted~\cite{xie2024recall,wang2025conrecalldetectingpretrainingdata} models, including GPT-2~\citep{radford2019language}, GPT-J-6B~\citep{gpt-j}, Llama-3.2-3B~\cite{dubey2024llama}, and a state-space model, Mamba-1.4B~\cite{gu2023mamba}. For each experiment, $\mathcal{M}^{\text{T}}$ is created by fine-tuning the corresponding pre-trained model on a specific dataset. The original pre-trained model serves as $\mathcal{M}^{\text{R}}$. Detailed fine-tuning hyperparameters and computational infrastructure are described in Appendix~\ref{app:training_details}.

\mypara{Attack Baselines} We evaluate our proposed method against a comprehensive suite of thirteen established MIA baselines, covering both reference-free and reference-based approaches. The reference-free attacks include the straightforward Loss score (average negative log-likelihood)~\cite{yeom2018}, methods based on input perturbations like ZLIB (which measures text compressibility) and Lowercase (which measures loss change after case modification)~\cite{carlini2021extracting}, tail-end distribution methods including Min-K\%~\cite{shi2024} and Min-K\%++~\cite{zhang2025}, and distribution-based DC-PDD~\cite{zhang2024pretraining}. The reference-based attacks include the foundational Ratio and Difference methods that compare the target model's average loss to a reference model's~\cite{watson2021importance}; context-aware query methods like ReCall~\cite{xie2024recall} and CON-ReCall~\cite{wang2025conrecalldetectingpretrainingdata} that test a model's ability to complete a prefix; self-prompt verification with SPV-MIA~\cite{fu2024}; and classifier-based attacks including a data-oriented Bag-of-Words baseline~\cite{das2025blind} and an Ensemble attack that uses multiple loss-based statistics as features~\cite{zhang2025softselectivedataobfuscation}. Our proposed method, \ourmethod, is evaluated using its ensemble configuration with $w_{\min}=2$ and $w_{\max}=40$. Implementation specifics and hyperparameter configurations for all baselines are detailed in Appendix~\ref{app:attack_imple}.

\mypara{Evaluation Metrics} We evaluate attack performance using AUC-ROC, TPR at low FPR thresholds (10\%, 1\%, 0.1\%), and model utility via perplexity shown in Appendix \ref{app:utility}. All metrics are reported with mean and standard deviation over 100 bootstrap runs following~\cite{bertail2008bootstrapping}. 

\subsection{Main Results}
\label{subsec:main_results}


The main attack performance results, presented in Table \ref{tab:main_results_part_one}, demonstrate that \ourmethod decisively outperforms all baselines across the six Cosmopedia datasets shown; results for the five real-world datasets are provided in Appendix~\ref{app:additional_results} due to space constraints, with consistently strong performance across all eleven datasets. \ourmethod achieves an average AUC of 0.826 compared to the strongest baseline, Ratio's 0.754, with consistent improvements across diverse domains; for instance, on the Stanford dataset, \ourmethod reaches an AUC of 0.845, significantly surpassing the 0.781 achieved by Ratio. The superiority of our method is most pronounced in the critical low-FPR regime. On the Web Samples v2 dataset, \ourmethod achieves a True Positive Rate at 1\% False Positive Rate (TPR@1\%FPR) of 19.8\%, more than doubling the 9.4\% from the strongest baseline on that metric, Difference. At the extreme 0.1\% FPR level on Khan Academy, \ourmethod identifies 2.6\% of members—a 3.7-fold increase over the next best method, Ensemble (0.7\%). This performance contrasts with reference-free baselines, whose near-random performance (AUC $\approx$ 0.6) validates that a reference model is essential for effectively attacking modern fine-tuned LLMs.
\subsection{Ablation Studies}
\label{subsec:ablation}

We conduct systematic ablation studies to validate each component of \ourmethod and quantify their individual contributions. 

\subsubsection{Optimal Window Size Determination}
\label{subsubsec:window_size}
\begin{figure}
    \centering
    \includegraphics[width=\linewidth]{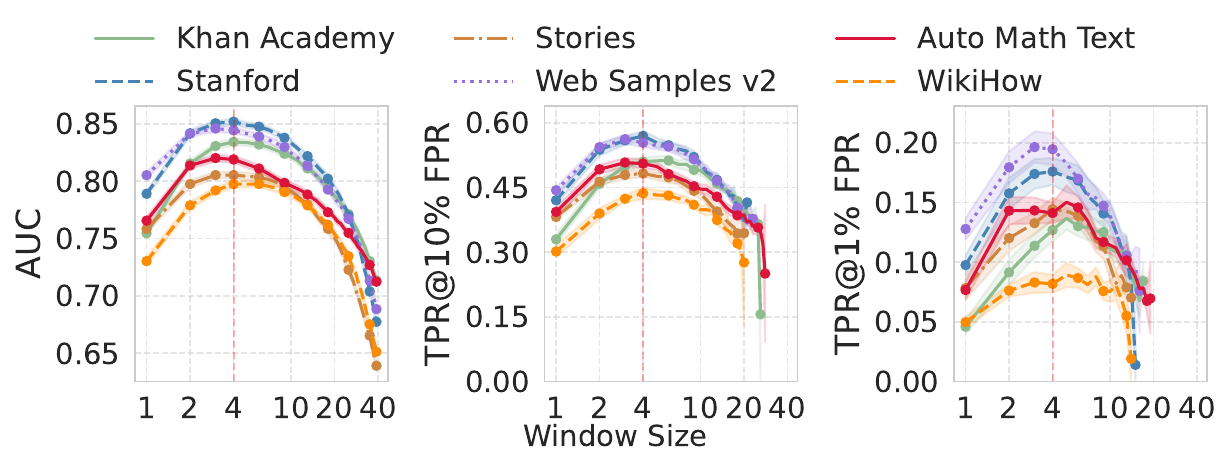}
    \caption{\textbf{Window size trade-off.} Performance of single-window \ourmethod attacks as a function of window size $w \in [1, 39]$ on six datasets. Metrics include AUC, TPR@10\% FPR, and TPR@1\% FPR. Shaded regions show the standard deviation over 100 bootstrap samples.}
    \label{fig:window_tradeoff_ablation}
\end{figure}

Figure~\ref{fig:window_tradeoff_ablation} empirically confirms our theoretical window size trade-off. Performance generally peaks at small windows ($w \in [3, 4]$) before degrading, reflecting the balance between accumulating sufficient signal and avoiding dilution. The exact optimum varies by dataset, e.g., Khan Academy peaks at $w=4$ (AUC 0.834) while Stories optimizes at $w=3$ (AUC 0.806), driven by unobservable parameters like signal sparsity $\rho_\delta$ and rare token frequency $\rho_\xi$.

Performance declines with larger windows; on Khan Academy, increasing $w$ from 4 to 32 drops AUC to 0.702, with TPR@1\%FPR collapsing to near zero beyond $w=20$. This validates our model: larger windows increase contamination probability and reduce the effective sample size (e.g., from 509 tests at $w=4$ to 257 at $w=256$), overwhelming the sparse membership signals.

\subsubsection{Ensemble Composition Analysis}
\label{subsubsec:ensemble}

\definecolor{low}{RGB}{255,245,240}
\definecolor{mid}{RGB}{252,146,114}
\definecolor{high}{RGB}{215,48,39}

\newcommand{\heatA}[1]{\cellcolor{high!75!white}#1}  
\newcommand{\heatB}[1]{\cellcolor{high!55!white}#1}  
\newcommand{\heatC}[1]{\cellcolor{high!35!white}#1}  
\newcommand{\heatD}[1]{\cellcolor{high!20!white}#1}  
\newcommand{\heatE}[1]{\cellcolor{high!8!white}#1}   

\begin{table}[t]
\centering
\caption{\textbf{Ensemble configuration comparison on Khan Academy dataset.} Performance of different window size combinations. The full ensemble achieves optimal balance between coverage and computational cost.}
\label{tab:ensemble_configs}
\small
\renewcommand{\arraystretch}{1.05}
\setlength{\tabcolsep}{3pt}
\begin{tabular}{@{}l@{\hspace{6pt}}c@{\hspace{5pt}}c@{\hspace{5pt}}c@{}}
\toprule
\textbf{Configuration} & \textbf{AUC} & \textbf{TPR@10\%} & \textbf{TPR@1\%} \\
\midrule
Single Best ($w$=4) & \heatB{0.8341\textsubscript{$\pm$.0024}} & \heatC{0.5103\textsubscript{$\pm$.0105}} & \heatD{0.1262\textsubscript{$\pm$.0094}} \\
Small Range & \heatB{0.8340\textsubscript{$\pm$.0028}} & \heatB{0.5195\textsubscript{$\pm$.0091}} & \heatD{0.1296\textsubscript{$\pm$.0090}} \\
Large Range & \heatE{0.7836\textsubscript{$\pm$.0030}} & \heatE{0.4293\textsubscript{$\pm$.0079}} & \heatE{0.0011\textsubscript{$\pm$.0110}} \\
Full Ensemble & \heatA{0.8369\textsubscript{$\pm$.0027}} & \heatA{0.5384\textsubscript{$\pm$.0076}} & \heatA{0.1464\textsubscript{$\pm$.0088}} \\
Extended & \heatC{0.8268\textsubscript{$\pm$.0030}} & \heatB{0.5209\textsubscript{$\pm$.0084}} & \heatC{0.1378\textsubscript{$\pm$.0085}} \\
Linear Spacing & \heatD{0.8241\textsubscript{$\pm$.0034}} & \heatC{0.5147\textsubscript{$\pm$.0096}} & \heatC{0.1389\textsubscript{$\pm$.0087}} \\
Random Selection & \heatD{0.8065\textsubscript{$\pm$.0026}} & \heatD{0.4720\textsubscript{$\pm$.0090}} & \heatD{0.1325\textsubscript{$\pm$.0078}} \\
\bottomrule
\end{tabular}
\end{table}

We evaluate different ensemble strategies for aggregating evidence across scales. Table~\ref{tab:ensemble_configs} compares seven configurations on Khan Academy, with consistent trends observed across all datasets shown in Appendix~\ref{app:additional_ensemble}. Our geometric ensemble follows the progression defined in Equation~\ref{eq:window_progression} with $w_{\min}=2$, $w_{\max}=40$, and $|W|=10$. This achieves the highest AUC, outperforming all alternatives.

Systematic baselines includes: \emph{Linear spacing} with $W = \{w_{\min} + i\Delta : i \in [0, |W|-1]\}$ where $\Delta = (w_{\max}-w_{\min})/(|W|-1)$ reaching AUC 0.8241, a 1.5\% drop. \emph{Extended coverage} using $W = \{w : w = w_{\min} + ki, k \in \mathbb{N}, w < n/2\}$ with larger stride $i=16$ reaches windows near $n/2$ but yields only AUC 0.8268. Range-restricted ensembles isolate scales: \emph{Small range} $W = [w_{\min}, w_{\min}+4]$ preserves 99.7\% of full performance, whereas \emph{Large range} $W = {w \in [18,50]: |w-w'|\geq7}$ falls to 0.7836 AUC with near-zero TPR@1\%FPR, confirming that large windows add noise rather than signal. Notably, most reasonable configurations achieve AUC above 0.80, demonstrating the robustness of window-based aggregation—even random selection of 10 windows yields AUC 0.8065. However, the geometric ensemble's systematic design provides crucial advantages in high-precision regimes: while AUC improves modestly (0.3\%), TPR@1\%FPR increases by 16\% over the single best window. This disproportionate gain in low-FPR detection, where security applications operate, justifies the minimal computational overhead (~6\% additional cost) and establishes geometric spacing as the optimal configuration for practical deployment. The consistent superiority across datasets confirms that geometric progression effectively balances the exploration-exploitation trade-off without requiring dataset-specific tuning.
\subsubsection{Aggregation Method Comparison}
\label{subsubsec:aggregation}
\begin{figure}[h]
    \centering
    \includegraphics[width=\linewidth]{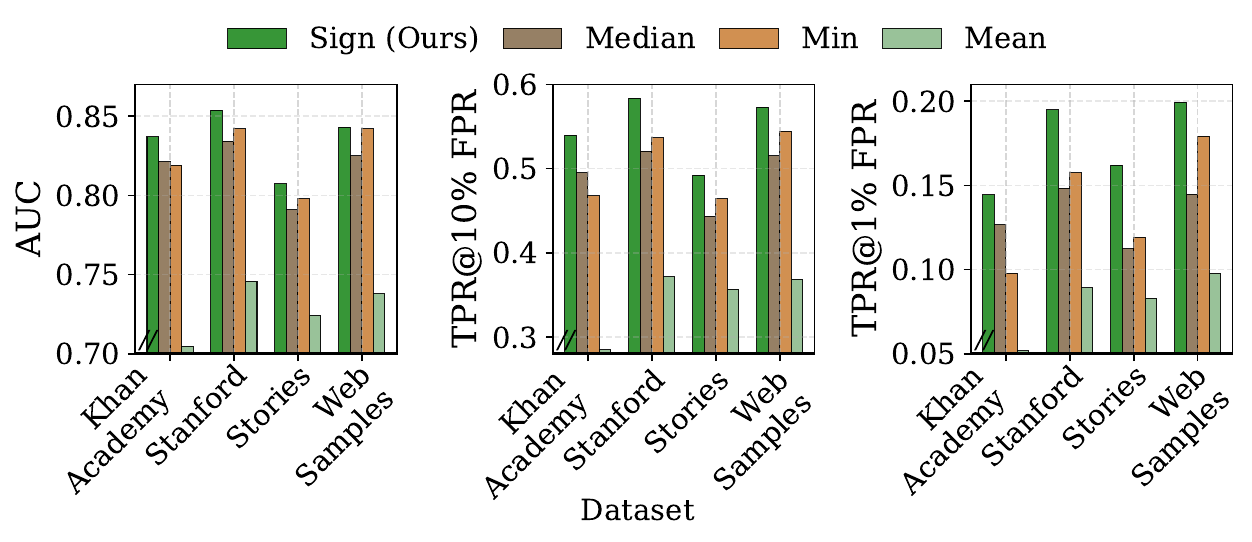}
    \caption{\textbf{Aggregation method comparison across datasets.} (a) AUC performance across four aggregation strategies. (b,c) TPR at low FPR thresholds shows amplified advantages in high-precision regimes.}
    \label{fig:aggregation}
\end{figure}

Figure~\ref{fig:aggregation} validates our theoretical prediction that sign-based aggregation provides superior robustness to long-tailed loss distributions. Sign aggregation achieves the highest average AUC (0.835) compared to mean (0.728), median (0.818), and min (0.825), with a 2.2$\times$ advantage in TPR@1\%FPR.




The relative performance of magnitude-based methods (mean, median, min) varies by dataset, reflecting different noise characteristics. However, all underperform sign aggregation, which achieves provable robustness through its bounded [0,1] output range and invariance to outlier magnitudes. Min aggregation's strong performance suggests membership manifests as consistently lower losses across multiple windows rather than isolated memorized passages. 
\subsubsection{Text Length Scaling}
\label{subsubsec:text_length}
\begin{figure}[h]
\centering
\includegraphics[width=\columnwidth]{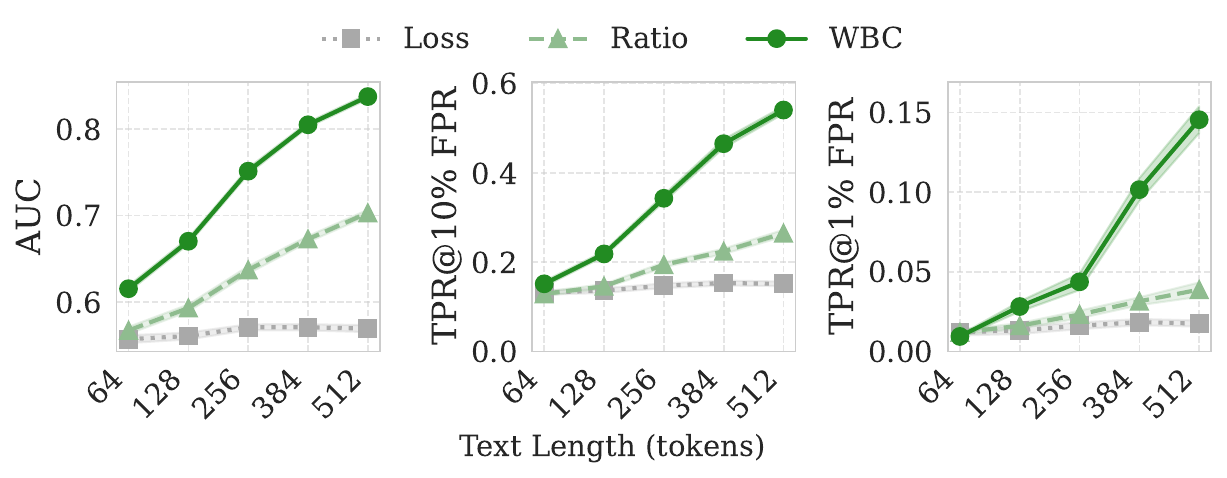}
\caption{\textbf{Attack performance scaling with text length.} Loss, Ratio, and \ourmethod on texts truncated to different length. (a) \ourmethod’s margin over baselines grows with $L$, reaching 20.8\% over Ratio at 512 tokens. (b,c) TPR scales super-linearly for \ourmethod.} 
\label{fig:text_length}
\end{figure}

Window-based analysis requires sufficient text to extract multiple measurements. Figure~\ref{fig:text_length} examines performance scaling by truncating 512-token samples to lengths $L \in \{64, 128, 256, 384, 512\}$. \ourmethod's advantage over baselines increases monotonically with length. At $L=64$ (permitting only 62 windows of size 3), \ourmethod achieves AUC 0.606 versus Ratio's 0.544—an 11.4\% improvement. This gap widens to 15.4\% at $L=256$ and 20.8\% at $L=512$. The scaling is super-linear for precision metrics: TPR@1\%FPR grows from 0.019 to 0.135 for \ourmethod (7.1$\times$ increase) versus 0.009 to 0.037 for Ratio (4.1$\times$ increase). The Loss baseline shows negligible length dependence, maintaining AUC $\approx 0.55$ across all lengths, confirming that reference comparison is essential regardless of text length. The relative gain of \ourmethod over Ratio—measured as $(AUC_{\text{\ourmethod}} - AUC_{\text{Ratio}})/AUC_{\text{Ratio}}$—increases from 11.4\% to 20.8\% as length grows, demonstrating that our voting mechanism's statistical power scales with the number of windows. Performance plateaus beyond $L=384$: extending to 512 tokens improves AUC by only 0.9\% (0.833 vs 0.825). 
\subsubsection{Component Importance Quantification}
\label{subsubsec:components}

Table~\ref{tab:ablation} quantifies each component's contribution. The reference model comparison proves essential; without it, AUC drops to 0.569 (near random), confirming that membership signals emerge from differential behavior between fine-tuned and base models rather than absolute confidence.  While design choices have modest individual impacts—ensemble aggregation (+0.3\%), geometric spacing (+2.6\%), sliding windows (+3.9\%)—the sign-based scoring proves critical with a 15.8\% performance drop when removed. More importantly, these components dramatically affect high-precision detection: sliding windows alone account for 51.2\% of TPR@1\%FPR (0.0715 vs 0.1464), demonstrating that maximizing measurement count is crucial for confident membership identification. 

\begin{table}[t]
\centering
\caption{\textbf{Component ablation study on Khan Academy dataset.}  Removal of \ourmethod components reveals reference model comparison and sign-based scoring as critical, contributing 32.0\% and 15.9\% of total performance, respectively.}
\label{tab:ablation}
\footnotesize
\renewcommand{\arraystretch}{1.08}
\setlength{\tabcolsep}{3.5pt}
\begin{tabular}{@{}l@{\hspace{8pt}}c@{\hspace{6pt}}c@{\hspace{8pt}}c@{\hspace{8pt}}c@{}}
\toprule
\textbf{Configuration} & \textbf{AUC} & \textbf{$\Delta$} & \textbf{TPR@10\%} & \textbf{TPR@1\%} \\
\midrule
\textbf{Full \ourmethod} & \textbf{0.8369}\textsubscript{$\pm$.0027} & — & \textbf{0.5384}\textsubscript{$\pm$.0076} & \textbf{0.1464}\textsubscript{$\pm$.0088} \\
\cmidrule(l){1-5}
\textit{Design choices} & & & & \\
~~w/o ensemble & 0.8341\textsubscript{$\pm$.0026} & \textcolor{gray}{$-$0.3\%} & 0.5093\textsubscript{$\pm$.0092} & 0.1278\textsubscript{$\pm$.0091} \\
~~w/o geometric & 0.8152\textsubscript{$\pm$.0029} & \textcolor{gray}{$-$2.6\%} & 0.5023\textsubscript{$\pm$.0088} & 0.1373\textsubscript{$\pm$.0080} \\
~~w/o sliding & 0.8042\textsubscript{$\pm$.0030} & \textcolor{gray}{$-$3.9\%} & 0.4195\textsubscript{$\pm$.0062} & 0.0715\textsubscript{$\pm$.0099} \\
\cmidrule(l){1-5}
\textit{Core components} & & & & \\
~~w/o sign-based & 0.7045\textsubscript{$\pm$.0038} & \textcolor{red!70!black}{$-$15.8\%} & 0.2835\textsubscript{$\pm$.0077} & 0.0508\textsubscript{$\pm$.0046} \\
~~w/o reference & 0.5693\textsubscript{$\pm$.0040} & \textcolor{red!70!black}{$-$31.9\%} & 0.1510\textsubscript{$\pm$.0053} & 0.0175\textsubscript{$\pm$.0022} \\
\bottomrule
\end{tabular}
\end{table}

\subsection{Generalization Across Model Scales and Architectures}
We test its generalizability across two key dimensions: model scale and architectural family. In these experiments, we compare \ourmethod against the representative baseline, Ratio, to provide a consistent reference point and evaluate the relative performance gap in new contexts.

\subsubsection{Performance Across Model Scales}

\begin{figure}[h]
    \centering
    \includegraphics[width=\linewidth]{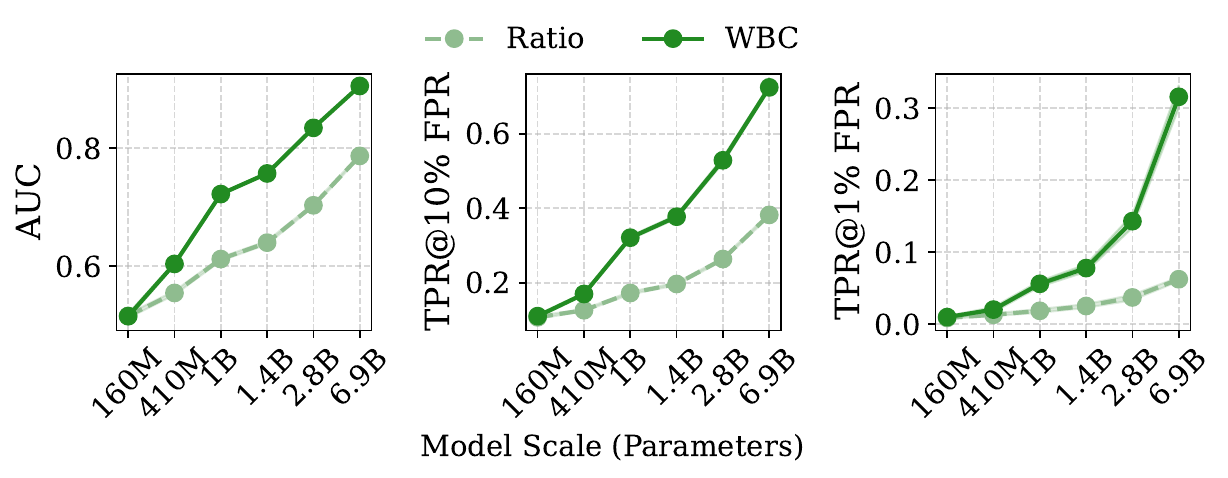}
    \caption{\textbf{Performance comparison of \ourmethod and Ratio methods across model scales.} Metrics include AUC, TPR@10\% FPR, and TPR@1\% FPR. Shaded regions denote standard deviations.}
    \label{fig:scale_generalization}
\end{figure}

The vulnerability to our window-based attack intensifies dramatically with model scale. Figure~\ref{fig:scale_generalization} shows attack performance on the Pythia suite from 160M to 6.9B parameters, revealing that vulnerability to window-based attacks increases dramatically with model scale. While both \ourmethod and Ratio perform near randomly at 160M parameters (AUC $\approx$ 0.51), their performance diverges sharply as model size grows.

At 2.8B parameters, \ourmethod achieves AUC 0.835 versus Ratio's 0.703. The advantage is most pronounced in high-confidence detection: TPR@1\%FPR reaches 14.3\% for \ourmethod versus 3.7\% for Ratio, nearly a 4$\times$ improvement. This widening gap reflects the fact that larger models possess greater memorization capacity, thereby exposing increased vulnerability to MIA. While global averaging dilutes these sharp local patterns, causing Ratio's effectiveness to plateau, \ourmethod's window-based detection captures them directly, with performance scaling in tandem with model capacity.

\subsubsection{Performance Across Model Architectures}

\begin{table}[t]
    \centering
    \footnotesize
    \scalefont{0.9}
    \setlength{\tabcolsep}{3pt}
    \caption{\textbf{Model performance comparison on Khan Academy dataset.} \ourmethod consistently outperforms the Ratio baseline across all architectures and scales.}
    \label{tab:khanacademy_results}
    \begin{tabular}{@{} l l c c c c @{}}
        \toprule
        \multirow{2}{*}{\textbf{Model}} & \multirow{2}{*}{\textbf{Method}} & \multicolumn{4}{c}{\textbf{Attack Performance}} \\
        \cmidrule(lr){3-6}
        & & \textbf{AUC} & \textbf{TPR@10\%} & \textbf{TPR@1\%} & \textbf{TPR@0.1\%} \\
        \midrule
        \multirow{2}{*}{GPT-2~\cite{radford2019language}} 
            & Ratio & 0.505\textsubscript{$\pm$.004} & 0.101\textsubscript{$\pm$.003} & 0.009\textsubscript{$\pm$.001} & 0.001\textsubscript{$\pm$.001} \\
            & \cellcolor{gray!15}\textbf{\ourmethod} & \cellcolor{gray!15}\textbf{0.505}\textsubscript{$\pm$.003} & \cellcolor{gray!15}\textbf{0.104}\textsubscript{$\pm$.003} & \cellcolor{gray!15}\textbf{0.013}\textsubscript{$\pm$.002} & \cellcolor{gray!15}\textbf{0.001}\textsubscript{$\pm$.001} \\
        \midrule
        \multirow{2}{*}{Mamba-1.4B~\cite{gu2023mamba}} 
            & Ratio & 0.592\textsubscript{$\pm$.003} & 0.154\textsubscript{$\pm$.005} & 0.018\textsubscript{$\pm$.002} & 0.001\textsubscript{$\pm$.000} \\
            & \cellcolor{gray!15}\textbf{\ourmethod} & \cellcolor{gray!15}\textbf{0.673}\textsubscript{$\pm$.002} & \cellcolor{gray!15}\textbf{0.274}\textsubscript{$\pm$.004} & \cellcolor{gray!15}\textbf{0.048}\textsubscript{$\pm$.003} & \cellcolor{gray!15}\textbf{0.004}\textsubscript{$\pm$.001} \\
        \midrule
        \multirow{2}{*}{Pythia-2.8B~\cite{biderman2023pythia}} 
            & Ratio & 0.703\textsubscript{$\pm$.003} & 0.264\textsubscript{$\pm$.004} & 0.037\textsubscript{$\pm$.004} & 0.003\textsubscript{$\pm$.001} \\
            & \cellcolor{gray!15}\textbf{\ourmethod} & \cellcolor{gray!15}\textbf{0.833}\textsubscript{$\pm$.002} & \cellcolor{gray!15}\textbf{0.529}\textsubscript{$\pm$.009} & \cellcolor{gray!15}\textbf{0.135}\textsubscript{$\pm$.008} & \cellcolor{gray!15}\textbf{0.026}\textsubscript{$\pm$.009} \\
        \midrule
        \multirow{2}{*}{Llama-3.2-3B~\cite{dubey2024llama}} 
            & Ratio & 0.861\textsubscript{$\pm$.003} & 0.551\textsubscript{$\pm$.010} & 0.135\textsubscript{$\pm$.014} & 0.019\textsubscript{$\pm$.006} \\
            & \cellcolor{gray!15}\textbf{\ourmethod} & \cellcolor{gray!15}\textbf{0.942}\textsubscript{$\pm$.002} & \cellcolor{gray!15}\textbf{0.831}\textsubscript{$\pm$.006} & \cellcolor{gray!15}\textbf{0.467}\textsubscript{$\pm$.015} & \cellcolor{gray!15}\textbf{0.193}\textsubscript{$\pm$.028} \\
        \midrule
        \multirow{2}{*}{GPT-J-6B~\cite{gpt-j}} 
            & Ratio & 0.868\textsubscript{$\pm$.002} & 0.564\textsubscript{$\pm$.009} & 0.126\textsubscript{$\pm$.007} & 0.013\textsubscript{$\pm$.005} \\
            & \cellcolor{gray!15}\textbf{\ourmethod} & \cellcolor{gray!15}\textbf{0.961}\textsubscript{$\pm$.001} & \cellcolor{gray!15}\textbf{0.887}\textsubscript{$\pm$.003} & \cellcolor{gray!15}\textbf{0.566}\textsubscript{$\pm$.016} & \cellcolor{gray!15}\textbf{0.209}\textsubscript{$\pm$.043} \\
        \bottomrule
    \end{tabular}
\end{table}

To ensure our findings are not specific to the Pythia architecture, Table \ref{tab:khanacademy_results} demonstrates \ourmethod's generalizability across diverse architectures on the Khan Academy dataset. On large-scale transformers, \ourmethod substantially outperforms Ratio: GPT-J-6B achieves AUC 0.961 versus 0.868, with TPR@1\%FPR of 56.6\% versus 12.6\% (4.5$\times$ improvement); Llama-3.2-3B shows similar gains with TPR@1\%FPR of 46.7\% versus 13.5\% (3.4$\times$ improvement). The advantage extends beyond transformers—on the state-space model Mamba-1.4B, \ourmethod maintains superiority (AUC 0.673 vs. 0.592), confirming that local signal aggregation is architecturally agnostic. Only GPT-2 shows near-random performance for both methods, consistent with minimal memorization in smaller models on complex datasets. These results establish that window-based detection exploits fundamental memorization patterns that transcend specific architectural choices.
\subsection{Robustness to Misaligned References}
\label{app:misaligned_references}
\begin{table}[t]
    \centering
    \footnotesize
    \scalefont{0.9}
    \setlength{\tabcolsep}{3pt}
    \caption{\textbf{Impact of Reference Model Mismatch.} We attack a Pythia-2.8B target using reference models with different sizes and different architectures.}
    \label{tab:misaligned_results}
    \begin{tabular}{@{} l l c c c c @{}}
        \toprule
        \multirow{2}{*}{\textbf{Ref. Model}} & \multirow{2}{*}{\textbf{Method}} & \multicolumn{4}{c}{\textbf{Khan Academy (Target: Pythia-2.8B)}} \\
        \cmidrule(lr){3-6}
        & & \textbf{AUC} & \textbf{TPR@10\%} & \textbf{TPR@1\%} & \textbf{TPR@0.1\%} \\
        \midrule
        \multirow{2}{*}{Pythia-160m} 
            & Ratio & 0.656\textsubscript{$\pm$.003} & 0.205\textsubscript{$\pm$.005} & 0.029\textsubscript{$\pm$.003} & 0.001\textsubscript{$\pm$.001} \\
            & \cellcolor{gray!15}\textbf{\ourmethod} & \cellcolor{gray!15}\textbf{0.692}\textsubscript{$\pm$.002} & \cellcolor{gray!15}\textbf{0.294}\textsubscript{$\pm$.005} & \cellcolor{gray!15}\textbf{0.062}\textsubscript{$\pm$.005} & \cellcolor{gray!15}\textbf{0.011}\textsubscript{$\pm$.004} \\
        \midrule
        \multirow{2}{*}{Pythia-1.4B} 
            & Ratio & 0.692\textsubscript{$\pm$.003} & 0.245\textsubscript{$\pm$.007} & 0.034\textsubscript{$\pm$.003} & 0.003\textsubscript{$\pm$.001} \\
            & \cellcolor{gray!15}\textbf{\ourmethod} & \cellcolor{gray!15}\textbf{0.774}\textsubscript{$\pm$.002} & \cellcolor{gray!15}\textbf{0.407}\textsubscript{$\pm$.008} & \cellcolor{gray!15}\textbf{0.077}\textsubscript{$\pm$.004} & \cellcolor{gray!15}\textbf{0.016}\textsubscript{$\pm$.006} \\
        \midrule
        \multirow{2}{*}{Pythia-6.9B} 
            & Ratio & 0.695\textsubscript{$\pm$.002} & 0.255\textsubscript{$\pm$.003} & 0.037\textsubscript{$\pm$.002} & 0.002\textsubscript{$\pm$.001} \\
            & \cellcolor{gray!15}\textbf{\ourmethod} & \cellcolor{gray!15}\textbf{0.769}\textsubscript{$\pm$.002} & \cellcolor{gray!15}\textbf{0.369}\textsubscript{$\pm$.008} & \cellcolor{gray!15}\textbf{0.071}\textsubscript{$\pm$.006} & \cellcolor{gray!15}\textbf{0.014}\textsubscript{$\pm$.004} \\
        \midrule
        \multirow{2}{*}{GPT-J-6B} 
            & Ratio & \textbf{0.685}\textsubscript{$\pm$.003} & 0.256\textsubscript{$\pm$.003} & 0.038\textsubscript{$\pm$.003} & 0.002\textsubscript{$\pm$.001} \\
            & \cellcolor{gray!15}\textbf{\ourmethod} & \cellcolor{gray!15}0.673\textsubscript{$\pm$.003} & \cellcolor{gray!15}\textbf{0.269}\textsubscript{$\pm$.006} & \cellcolor{gray!15}\textbf{0.044}\textsubscript{$\pm$.003} & \cellcolor{gray!15}\textbf{0.003}\textsubscript{$\pm$.002} \\
        \bottomrule
    \end{tabular}
\end{table}

We simulate a scenario where the adversary does not have access to the exact base model used for fine-tuning. We fix the target model as Pythia-2.8B and evaluate performance using distinct reference models with size mismatch (Pythia-160m, Pythia-1.4B, and Pythia-6.9B) and architecture mismatch (GPT-J-6B).

The results are summarized in \Cref{tab:misaligned_results}. When using Pythia-1.4B (a model half the size of the target) as a reference, \ourmethod achieves an AUC of 0.774 and a TPR of 7.7\% at 1\% FPR, significantly outperforming the Ratio baseline (AUC 0.692, TPR 3.4\%). With a different architecture (GPT-J), \ourmethod maintains a performance advantage in the critical low false-positive regime, though it trails slightly in overall AUC (0.673 vs 0.685). This AUC deficit likely stems from tokenization and architectural disparities, introducing noise into local comparisons that global averaging smooths out.
\subsection{Defense Evaluation}

Given that \ourmethod is highly effective, it is important to assess whether privacy-preserving training techniques can defend against it. We evaluate \ourmethod against three defense mechanisms spanning different protection strategies: differential privacy, parameter-efficient training, and data obfuscation. 

\subsubsection{Differential Privacy}
\begin{table}[t]
    \centering
    \footnotesize
    \scalefont{0.9}
    \setlength{\tabcolsep}{3pt}
    \caption{\textbf{Attack performance under differential privacy on Khan Academy dataset.} Evaluation at privacy budgets $\epsilon \in \{1, 4, 8, \infty\}$ with fixed $\delta = 10^{-5}$.}
    \label{tab:dp_results}
    \begin{tabular}{@{} >{\centering\arraybackslash}p{1.2cm} l c c c c c @{}}
        \toprule
        \multirow{2}{*}{\textbf{$\epsilon$}} & \multirow{2}{*}{\textbf{Method}} & \multicolumn{4}{c}{\textbf{Attack Performance}} & \multirow{2}{*}{\textbf{PPL}} \\
        \cmidrule(lr){3-6}
        & & \textbf{AUC} & \textbf{TPR@10\%} & \textbf{TPR@1\%} & \textbf{TPR@0.1\%} & \\
        \midrule
        \multirow{2}{*}{$\infty$} 
            & Ratio & 0.703\textsubscript{$\pm$.003} & 0.264\textsubscript{$\pm$.004} & 0.037\textsubscript{$\pm$.004} & 0.003\textsubscript{$\pm$.001} & \multirow{2}{*}{3.49} \\
            & \cellcolor{gray!15}\textbf{\ourmethod} & \cellcolor{gray!15}\textbf{0.837}\textsubscript{$\pm$.003} & \cellcolor{gray!15}\textbf{0.538}\textsubscript{$\pm$.008} & \cellcolor{gray!15}\textbf{0.146}\textsubscript{$\pm$.008} & \cellcolor{gray!15}\textbf{0.026}\textsubscript{$\pm$.009} & \\
        \midrule
        \multirow{2}{*}{8} 
            & Ratio & 0.642\textsubscript{$\pm$.004} & 0.198\textsubscript{$\pm$.006} & 0.024\textsubscript{$\pm$.003} & 0.002\textsubscript{$\pm$.001} & \multirow{2}{*}{3.85} \\
            & \cellcolor{gray!15}\textbf{\ourmethod} & \cellcolor{gray!15}\textbf{0.751}\textsubscript{$\pm$.003} & \cellcolor{gray!15}\textbf{0.358}\textsubscript{$\pm$.008} & \cellcolor{gray!15}\textbf{0.078}\textsubscript{$\pm$.006} & \cellcolor{gray!15}\textbf{0.013}\textsubscript{$\pm$.003} & \\
        \midrule
        \multirow{2}{*}{4} 
            & Ratio & 0.591\textsubscript{$\pm$.005} & 0.147\textsubscript{$\pm$.005} & 0.015\textsubscript{$\pm$.002} & 0.001\textsubscript{$\pm$.000} & \multirow{2}{*}{4.68} \\
            & \cellcolor{gray!15}\textbf{\ourmethod} & \cellcolor{gray!15}\textbf{0.674}\textsubscript{$\pm$.004} & \cellcolor{gray!15}\textbf{0.254}\textsubscript{$\pm$.007} & \cellcolor{gray!15}\textbf{0.042}\textsubscript{$\pm$.005} & \cellcolor{gray!15}\textbf{0.006}\textsubscript{$\pm$.002} & \\
        \midrule
        \multirow{2}{*}{1} 
            & Ratio & 0.524\textsubscript{$\pm$.006} & 0.108\textsubscript{$\pm$.004} & 0.008\textsubscript{$\pm$.002} & 0.000\textsubscript{$\pm$.000} & \multirow{2}{*}{4.77} \\
            & \cellcolor{gray!15}\textbf{\ourmethod} & \cellcolor{gray!15}\textbf{0.561}\textsubscript{$\pm$.005} & \cellcolor{gray!15}\textbf{0.135}\textsubscript{$\pm$.005} & \cellcolor{gray!15}\textbf{0.019}\textsubscript{$\pm$.003} & \cellcolor{gray!15}\textbf{0.001}\textsubscript{$\pm$.001} & \\
        \bottomrule
    \end{tabular}
\end{table}

Table~\ref{tab:dp_results} reveals that while DP-SGD reduces absolute attack success, \ourmethod maintains substantial relative advantages across all privacy budgets. At moderate privacy ($\epsilon = 8$), \ourmethod achieves AUC 0.751 versus Ratio's 0.642, with a more pronounced 3.25$\times$ advantage in TPR@1\%FPR (0.078 vs 0.024). This gap persists at strong privacy ($\epsilon = 1$): while both methods approach random guessing, \ourmethod still achieves 2.4$\times$ higher TPR@1\%FPR. While global noise reduces overall signal strength, local coherence within text windows remains partially intact. Notably, stronger privacy guarantees ($\epsilon = 1$) provide limited additional protection—AUC decreases from $\epsilon = 8$ (0.751 to 0.561) while perplexity increases by 37\%. This modest utility cost makes DP-SGD more practical than previously thought, though window-based attacks remain partially effective even under strong privacy guarantees.

\subsubsection{Low-Rank Adaptation}
\begin{table}[t]
    \centering
    \footnotesize
    \scalefont{0.9}
    \setlength{\tabcolsep}{3pt}
    \caption{\textbf{Attack performance under LoRA on Khan Academy dataset.} Evaluation with ranks $r \in \{8, 16, 32, 64\}$ and scaling factor $\alpha = 2r$.}
    \label{tab:lora}
    \begin{tabular}{@{} >{\centering\arraybackslash}p{0.3cm} l c c c c c c @{}}
        \toprule
        \multirow{2}{*}{\textbf{$r$}} & \multirow{2}{*}{\textbf{Method}} & \multicolumn{4}{c}{\textbf{Attack Performance}} & \multirow{2}{*}{\textbf{Trn. \%}} & \multirow{2}{*}{\textbf{PPL}} \\
        \cmidrule(lr){3-6}
        & & \textbf{AUC} & \textbf{TPR@10\%} & \textbf{TPR@1\%} & \textbf{TPR@0.1\%} & & \\
        \midrule
        \multirow{2}{*}{$\infty$} 
            & Ratio & 0.703\textsubscript{$\pm$.003} & 0.264\textsubscript{$\pm$.004} & 0.037\textsubscript{$\pm$.004} & 0.003\textsubscript{$\pm$.001} & \multirow{2}{*}{100\%} & \multirow{2}{*}{3.49} \\
            & \cellcolor{gray!15}\textbf{\ourmethod} & \cellcolor{gray!15}\textbf{0.837}\textsubscript{$\pm$.003} & \cellcolor{gray!15}\textbf{0.538}\textsubscript{$\pm$.008} & \cellcolor{gray!15}\textbf{0.146}\textsubscript{$\pm$.008} & \cellcolor{gray!15}\textbf{0.026}\textsubscript{$\pm$.009} & & \\
        \midrule
        \multirow{2}{*}{64} 
            & Ratio & 0.615\textsubscript{$\pm$.003} & 0.186\textsubscript{$\pm$.005} & 0.022\textsubscript{$\pm$.002} & 0.002\textsubscript{$\pm$.001} & \multirow{2}{*}{0.75\%} & \multirow{2}{*}{3.77} \\
            & \cellcolor{gray!15}\textbf{\ourmethod} & \cellcolor{gray!15}\textbf{0.748}\textsubscript{$\pm$.002} & \cellcolor{gray!15}\textbf{0.378}\textsubscript{$\pm$.007} & \cellcolor{gray!15}\textbf{0.074}\textsubscript{$\pm$.005} & \cellcolor{gray!15}\textbf{0.009}\textsubscript{$\pm$.002} & & \\
        \midrule
        \multirow{2}{*}{32} 
            & Ratio & 0.592\textsubscript{$\pm$.003} & 0.164\textsubscript{$\pm$.004} & 0.018\textsubscript{$\pm$.002} & 0.001\textsubscript{$\pm$.000} & \multirow{2}{*}{0.38\%} & \multirow{2}{*}{3.82} \\
            & \cellcolor{gray!15}\textbf{\ourmethod} & \cellcolor{gray!15}\textbf{0.698}\textsubscript{$\pm$.002} & \cellcolor{gray!15}\textbf{0.312}\textsubscript{$\pm$.006} & \cellcolor{gray!15}\textbf{0.048}\textsubscript{$\pm$.004} & \cellcolor{gray!15}\textbf{0.005}\textsubscript{$\pm$.001} & & \\
        \midrule
        \multirow{2}{*}{16} 
            & Ratio & 0.564\textsubscript{$\pm$.003} & 0.142\textsubscript{$\pm$.004} & 0.014\textsubscript{$\pm$.001} & 0.001\textsubscript{$\pm$.000} & \multirow{2}{*}{0.19\%} & \multirow{2}{*}{3.88} \\
            & \cellcolor{gray!15}\textbf{\ourmethod} & \cellcolor{gray!15}\textbf{0.634}\textsubscript{$\pm$.002} & \cellcolor{gray!15}\textbf{0.234}\textsubscript{$\pm$.005} & \cellcolor{gray!15}\textbf{0.029}\textsubscript{$\pm$.003} & \cellcolor{gray!15}\textbf{0.003}\textsubscript{$\pm$.001} & & \\
        \midrule
        \multirow{2}{*}{8} 
            & Ratio & 0.541\textsubscript{$\pm$.004} & 0.124\textsubscript{$\pm$.003} & 0.011\textsubscript{$\pm$.002} & 0.000\textsubscript{$\pm$.000} & \multirow{2}{*}{0.09\%} & \multirow{2}{*}{3.93} \\
            & \cellcolor{gray!15}\textbf{\ourmethod} & \cellcolor{gray!15}\textbf{0.578}\textsubscript{$\pm$.003} & \cellcolor{gray!15}\textbf{0.171}\textsubscript{$\pm$.004} & \cellcolor{gray!15}\textbf{0.018}\textsubscript{$\pm$.002} & \cellcolor{gray!15}\textbf{0.002}\textsubscript{$\pm$.001} & & \\
        \bottomrule
    \end{tabular}
\end{table}

LoRA~\cite{hu2022lora}, while designed for parameter efficiency, provides unintended privacy benefits by constraining memorization capacity~\cite{zhang2025softselectivedataobfuscation,luo2024privacy,amit2024sok}. The defense mechanism stems from capacity constraints: low-rank factorizations $W = W_0 + BA$ with $r \ll \min(d, k)$ force the model to compress information into a rank-$r$ subspace, favoring generalizable patterns over sample-specific memorization. We note that this protection relies on accepting a slight utility decrease; recent work indicates LoRA is not a robust defense under strict utility matching~\cite{ran2025loraleak}. While both attacks degrade under LoRA, \ourmethod maintains a consistent advantage—2.67$\times$ higher TPR@1\%FPR at rank 32 (0.048 vs 0.018) compared to 3.95$\times$ for full fine-tuning. This persistence suggests that even constrained memorization retains localized patterns that window-based detection exploits. LoRA thus offers meaningful protection with minimal utility cost, though sophisticated attacks targeting local signals remain partially effective.

\subsubsection{Selective data Obfuscation in LLM Fine-Tuning}
\begin{table}[t]
    \centering
    \footnotesize
    \scalefont{0.9}
    \setlength{\tabcolsep}{3pt}
    \caption{\textbf{Attack performance against SOFT defense on Khan Academy dataset.} SOFT configured with selection ratio $\alpha=0.3$ and paraphrase ratio $\beta=0.5$.}
    \label{tab:soft_defense}
    \begin{tabular}{@{} >{\centering\arraybackslash}p{1.2cm} l c c c c c @{}}
        \toprule
        \multirow{2}{*}{\textbf{Defense}} & \multirow{2}{*}{\textbf{Method}} & \multicolumn{4}{c}{\textbf{Attack Performance}} & \multirow{2}{*}{\textbf{PPL}} \\
        \cmidrule(lr){3-6}
        & & \textbf{AUC} & \textbf{TPR@10\%} & \textbf{TPR@1\%} & \textbf{TPR@0.1\%} & \\
        \midrule
        \multirow{2}{*}{Baseline} 
            & Ratio & 0.703\textsubscript{$\pm$.003} & 0.264\textsubscript{$\pm$.004} & 0.037\textsubscript{$\pm$.004} & 0.003\textsubscript{$\pm$.001} & \multirow{2}{*}{3.49} \\
            & \cellcolor{gray!15}\textbf{\ourmethod} & \cellcolor{gray!15}\textbf{0.837}\textsubscript{$\pm$.003} & \cellcolor{gray!15}\textbf{0.538}\textsubscript{$\pm$.008} & \cellcolor{gray!15}\textbf{0.146}\textsubscript{$\pm$.008} & \cellcolor{gray!15}\textbf{0.026}\textsubscript{$\pm$.009} & \\
        \midrule
        \multirow{2}{*}{SOFT} 
            & Ratio & 0.494\textsubscript{$\pm$.003} & 0.100\textsubscript{$\pm$.004} & 0.010\textsubscript{$\pm$.001} & 0.001\textsubscript{$\pm$.000} & \multirow{2}{*}{3.48} \\
            & \cellcolor{gray!15}\textbf{\ourmethod} & \cellcolor{gray!15}\textbf{0.494}\textsubscript{$\pm$.004} & \cellcolor{gray!15}\textbf{0.100}\textsubscript{$\pm$.003} & \cellcolor{gray!15}\textbf{0.012}\textsubscript{$\pm$.002} & \cellcolor{gray!15}\textbf{0.001}\textsubscript{$\pm$.000} & \\
        \bottomrule
    \end{tabular}
\end{table}

SOFT~\cite{zhang2025softselectivedataobfuscation} defends against membership inference by selectively paraphrasing influential training samples identified through loss-based thresholds. Table~\ref{tab:soft_defense} shows SOFT's effectiveness: it reduces both \ourmethod and Ratio from strong detection (AUC 0.837 and 0.703, respectively) to near-random performance (AUC $\approx$ 0.494). At TPR@1\%FPR, \ourmethod's advantage over Ratio shrinks from 3.9$\times$ to marginal (0.012 vs 0.010), a 92\% reduction in detection capability. 


\section{Conclusion}


We demonstrated that membership inference attacks against fine-tuned LLMs fail as global averaging is fundamentally unsuitable for detecting sparse, extremal memorization events. Instead, membership signals manifest as rare, localized patterns that are masked by long-tailed noise from domain adaptation. We thus presented \ourmethod, a membership inference attack that replaces global averaging with localized window-based analysis. The method slides windows of varying sizes across token sequences, computes binary comparisons between reference and target model losses, and aggregates these local votes using sign statistics, achieving 2-3$\times$ higher detection rates than existing methods across eleven datasets. While our empirical results demonstrate the effectiveness of the sign test, the shift from global to localized analysis opens a rich design space. We note that \ourmethod represents one effective instance of this paradigm; given the sparse, spike-like structure of membership signals we identified, future work could explore alternative aggregation functions or statistical modeling techniques to further enhance detection.

\section{Ethical Considerations}

We conducted a comprehensive stakeholder-based ethics analysis following the Menlo Report principles. Our research develops a membership inference attack that achieves 2--3$\times$ higher detection rates than existing methods, raising important privacy considerations that we carefully evaluated throughout the research process.

\mypara{Stakeholders and Impacts} 
We identified three primary stakeholder groups: ML practitioners and organizations deploying fine-tuned LLMs who may face increased privacy risks from our more effective attack; individuals whose data appears in training datasets and may be exposed through successful membership inference; and the broader research community working to understand and mitigate privacy risks in machine learning systems. For practitioners, our attack demonstrates that current assumptions about membership inference difficulty significantly underestimate actual risks, particularly for localized memorization patterns. This knowledge enables more informed decisions about data handling and privacy-preserving training techniques. For individuals, while our attack could theoretically enable adversaries to determine if their data was used in training, we note that our experiments used only publicly available datasets and synthetic data, avoiding any direct privacy violations. The research community benefits from our theoretical insights into why localized analysis outperforms global averaging, enabling the development of more targeted defenses.

We implemented several measures to minimize potential harm. First, all experiments used publicly available models and datasets, eliminating risks to proprietary systems or private data. We specifically chose datasets like Cosmopedia (synthetic data), WikiText-103, and other public corpora where membership disclosure poses minimal privacy risk. Second, we evaluated multiple defense mechanisms, including differential privacy, SOFT, and LoRA, providing concrete guidance for practitioners seeking to protect their systems. Our results show that while these defenses reduce attack effectiveness, they do not eliminate the vulnerability, highlighting the need for continued research into privacy-preserving training methods. Third, we responsibly disclose the limitations of our attack, including its requirement for score-based access to both target and reference models, making it impractical for many real-world scenarios without legitimate access.

\mypara{Justification} 
We proceeded with publication after determining that the benefits outweigh the potential harms. Our attack serves as a diagnostic tool for quantifying existing privacy risks---it requires only score-based black-box access that legitimate users already possess, introducing no new attack surfaces. By revealing that localized memorization is more detectable than previously believed and simultaneously evaluating defenses, we enable practitioners to make informed privacy decisions rather than relying on false confidence in inadequate measures. Publication through peer review at a defensive security venue ensures our findings reach those developing countermeasures rather than malicious actors.

\section{Open Science}


The complete implementation of our Window-Based Comparison attack, including all baseline methods, training and evaluation scripts, is available at \url{https://github.com/Stry233/WBC} and archived at \url{https://doi.org/10.5281/zenodo.17968678}. The archive includes the core WBC attack implementation with configurable window sizes and aggregation methods, implementations of all thirteen baseline attacks used in our evaluation, scripts for fine-tuning models on the datasets used in our experiments, and evaluation metrics including AUC and TPR at various FPR thresholds. All code is properly documented with clear instructions for setup and execution.

Our experiments utilize publicly available datasets accessible through standard channels: Cosmopedia~\cite{benallal2024cosmopedia} subsets via HuggingFace~\cite{wolf2019huggingface}, WikiText-103~\cite{merity2016pointer}, XSum~\cite{narayan2018don}, Amazon Reviews~\cite{ni2019justifying}, CC-News~\cite{Hamborg2017}, and Reddit~\cite{reddit_title_body_dataset} datasets through their respective public repositories. We provide scripts to automatically download and preprocess these datasets into the format required for our experiments. For models, we use the publicly available Pythia suite (160M to 6.9B parameters)~\cite{biderman2023pythia}, GPT-2~\cite{brown2020language}, GPT-J-6B~\cite{gpt-j}, Llama-3.2-3B~\cite{dubey2024llama}, and Mamba-1.4B~\cite{gu2023mamba}, all accessible through HuggingFace~\cite{wolf2019huggingface}. Our repository includes model configuration files and fine-tuning scripts to reproduce the exact target models used in our evaluation.

Reproducing our results requires access to GPUs capable of running models up to 6.9B parameters. We provide guidance for scaling experiments to smaller models for researchers with limited computational resources. The complete experimental suite required approximately 500 GPU-hours on NVIDIA A100 GPUs, though individual experiments can be run with significantly fewer resources. 

\bibliographystyle{plain}
\bibliography{references}

\appendix
\section{Notation Summary}
\label{app:notation}
This section provides a summary of core mathematical notation used throughout the paper to ensure clarity and consistency. Table~\ref{tab:app_notation_summary} lists these symbols and their meanings in the context of the \ourmethod attack.

\begin{table*}
\centering
\caption{Summary of Core Notation.}
\label{tab:app_notation_summary}
\begin{tabularx}{\linewidth}{@{}p{3.5cm} >{\raggedright\arraybackslash}X@{}}
\toprule
\textbf{Symbol} & \textbf{Meaning} \\
\midrule
\multicolumn{2}{@{}l}{\textit{General Symbols}} \\
\cmidrule(r){1-2}
$\mathbf{x} = (x_1, \ldots, x_n)$ & Text sequence of $n$ tokens. \\
$n$ & Length of sequence $\mathbf{x}$. \\
$D_{\text{train}}$ & Fine-tuning dataset (member set). \\
$\mathcal{M}$ & Generic language model. \\
$\mathcal{M}^{\text{T}}$ & Target model (fine-tuned on $D_{\text{train}}$). \\
$\mathcal{M}^{\text{R}}$ & Reference model (pre-trained base model). \\
$\theta$ & Parameters of a language model. \\
$\hat{m}(\mathbf{x})$ & Predicted membership status ($1$ for member, $0$ for non-member). \\
$A(\cdot)$ & Attack function for membership inference. \\
\midrule
\multicolumn{2}{@{}l}{\textit{Loss and Probability Symbols}} \\
\cmidrule(r){1-2}
$p(\mathbf{x})$ & Joint probability of sequence $\mathbf{x}$. \\
$p_{\mathcal{M}}(x_i \mid x_1, \ldots, x_{i-1})$ & Conditional probability of token $x_i$ given prefix. \\
$\ell_i^{\mathcal{M}}$ & Per-token negative log-likelihood for model $\mathcal{M}$ at position $i$. \\
$\ell_j^{\text{T}}, \ell_j^{\text{R}}$ & Per-token losses for target and reference models. \\
$\Delta_j$ & Loss difference: $\ell_j^{\text{R}} - \ell_j^{\text{T}}$. \\
\midrule
\multicolumn{2}{@{}l}{\textit{Theoretical Model Symbols (Section~\ref{subsec:theory})}} \\
\cmidrule(r){1-2}
$\delta_j(\mathbf{x})$ & Membership signal at position $j$ (present only for members). \\
$\xi_j$ & Rare token effect causing extreme loss differences. \\
$\epsilon_j \sim \mathcal{N}(0, \sigma^2)$ & Baseline noise at position $j$. \\
$\rho_\delta$ & Probability of membership signal occurrence (sparsity). \\
$\rho_\xi$ & Probability of rare token occurrence. \\
$\gamma_j, \bar{\gamma}$ & Membership signal strength (individual and average). \\
$Y_j$ & Magnitude of rare token event. \\
$\sigma^2$ & Variance of baseline noise. \\
$\mathbb{I}[\cdot]$ & Indicator function. \\
\midrule
\multicolumn{2}{@{}l}{\textit{Window-Based Analysis Symbols}} \\
\cmidrule(r){1-2}
$w$ & Window size (number of consecutive tokens). \\
$W$ & Set of window sizes used in ensemble. \\
$w_{\min}, w_{\max}$ & Minimum and maximum window sizes in ensemble. \\
$S_i(w)$ & Windowed sum of loss differences starting at position $i$. \\
$T_{\text{mean}}$ & Mean-based aggregation statistic. \\
$T_{\text{sign}}$ & Sign-based aggregation statistic. \\
$T_{\text{sign}}(w)$ & Window-specific sign score for window size $w$. \\
$S_{\text{WBC}}$ & Final \ourmethod membership score. \\
$n - w + 1$ & Number of sliding windows for sequence of length $n$. \\
\bottomrule
\end{tabularx}
\end{table*}

\section{Optimal Window Size Analysis}

\label{app:window_optimization}

To formalize the challenge of selecting an optimal window size $w$, we analyze the sign test's detection capability. For a window of size $w$, let $p_w^{(1)}$ denote the probability that $S_i(w) > 0$ for members and $p_w^{(0)}$ for non-members. Under our model, $p_w^{(0)} = 0.5$ due to symmetric noise distribution, while:

\begin{equation}
p_w^{(1)} = \Phi\left(\frac{\rho_\delta w \bar{\gamma}}{\sqrt{w\sigma^2 + \rho_\xi w \mathbb{E}[Y^2]}}\right),
\end{equation}

where $\Phi$ is the standard normal CDF. The separation $p_w^{(1)} - 0.5$ determines the sign test's discriminative power.

The sign statistic $T_{\text{sign}}(w)$ counts the fraction of windows with positive sums. For non-members, $\mathbb{E}[T_{\text{sign}}(w)] = 0.5$, while for members, $\mathbb{E}[T_{\text{sign}}(w)] = p_w^{(1)}$. The detectability depends on both this separation and the variance of $T_{\text{sign}}(w)$.

Due to window overlap, adjacent windows exhibit strong positive correlation:

\begin{equation}
\text{Var}[T_{\text{sign}}(w)] = \frac{1}{(n-w+1)^2} \sum_{i,j} \text{Cov}[\mathbb{I}[S_i(w) > 0], \mathbb{I}[S_j(w) > 0]].
\end{equation}

The covariance structure depends on window overlap: windows $i$ and $j$ share $\max(0, w - |i-j|)$ tokens. For $|i-j| \geq w$, the windows are disjoint and approximately independent. Since each group of $w$ consecutive windows contains at most one independent observation, the effective sample size is approximately $n/w$.

For a proportion estimator with effective sample size $n_{\text{eff}} \approx n/w$:

\begin{equation}
\text{Var}[T_{\text{sign}}(w)] \approx \frac{p_w^{(1)}(1-p_w^{(1)})}{n/w} = \frac{w \cdot p_w^{(1)}(1-p_w^{(1)})}{n}
\end{equation}

The power of the sign test thus becomes:

\begin{equation}
\text{Power}(w) \propto \frac{(p_w^{(1)} - 0.5)^2}{\text{Var}[T_{\text{sign}}(w)]} \approx \frac{(p_w^{(1)} - 0.5)^2 \cdot n}{w \cdot p_w^{(1)}(1-p_w^{(1)})}.
\end{equation}

This reveals the fundamental trade-off: the numerator $(p_w^{(1)} - 0.5)^2$ increases with $w$ as more tokens improve signal accumulation, but the denominator contains $w$ explicitly, reflecting the loss of independent tests. Even if doubling $w$ increases $(p_w^{(1)} - 0.5)$ by 40\%, the power may still decrease by 30\% due to the linear penalty from reduced effective sample size.

The optimal window size $w^*$ that maximizes this power depends critically on the parameters $\rho_\delta$, $\rho_\xi$, $\bar{\gamma}$, and $\mathbb{E}[Y^2]$. These parameters are unknown and vary across datasets and even within documents. Attempting to estimate them creates a circular dependency: identifying signal-containing windows requires knowing the parameters, but parameter estimation requires identifying these windows. This makes finding a single optimal window size infeasible in practice, motivating our ensemble approach.

\section{Supplementary Results}

\subsection{Experimental Settings}
\label{app:experimental_settings}

This section provides additional implementation details and computational requirements for our experimental evaluation that complement the main text.

\subsubsection{Dataset Details}
\label{app:datasets}

Our evaluation spans two categories of datasets for comprehensive evaluation.

\mypara{Cosmopedia Datasets} We utilize seven subsets from Cosmopedia~\cite{benallal2024cosmopedia}, a large-scale synthetic dataset generated by Mixtral-8x7B-Instruct-v0.1~\cite{jiang2023mistral7b} containing over 25 billion tokens. The Web Samples v2 split, which constitutes part of the largest portion of Cosmopedia, contains samples from an internal web dataset similar to RefinedWeb~\cite{penedo2023refinedweb}, with refined prompts requesting more depth in concept explanations. The Stanford split uses scraped course outlines from stanford.edu, prompting the model with individual course units. The Stories split adds commonsense and day-to-day knowledge by leveraging samples from UltraChat~\cite{ding2023enhancing} and OpenHermes2.5, synthetic instruction-tuning datasets covering diverse topics. For Khan Academy, we use scraped course outlines from KhanAcademy.org to generate textbook-style content. The AutoMathText split improves science knowledge using samples from the AutoMathText~\cite{zhang2024autonomous} dataset as seeds, covering mathematics and other scientific domains. Finally, the wikiHow split generates articles from scraped WikiHow~\cite{koupaee2018wikihow} titles, providing procedural and instructional content.

\mypara{Real-World Benchmarks} We complement the synthetic data with five established text benchmarks from prior work~\cite{fu2024}. WikiText-103~\cite{merity2016pointer} provides Wikipedia articles with long-term dependencies, while XSum~\cite{narayan2018don} contains BBC news articles paired with abstractive summaries. Amazon Reviews~\cite{ni2019justifying} offers product reviews across multiple categories, CC-News~\cite{Hamborg2017} includes news articles crawled from Common Crawl, and Reddit~\cite{reddit_title_body_dataset} provides social media posts with titles and bodies. All datasets use balanced 10,000-sample member/non-member splits filtered to minimum 512 tokens to ensure sufficient context for window-based analysis.

\subsubsection{Dataset Construction and Preprocessing}
\label{app:dataset_const}
For dataset preparation, we implement a unified preprocessing pipeline that handles both synthetic Cosmopedia subsets~\cite{benallal2024cosmopedia} and real-world benchmarks. Each dataset undergoes tokenization using the Pythia tokenizer with a strict 512-token length requirement. The preprocessing employs multiprocessing across CPU cores to efficiently filter samples below the minimum length threshold before constructing balanced 10,000-sample member and non-member splits. Texts exceeding 512 tokens are truncated at exact token boundaries to avoid partial word artifacts. The random sampling uses a fixed seed (42) for reproducibility, with member samples drawn from the training split and non-members from held-out data of the same distribution.

\subsubsection{Infrastructure and Training Details}
\label{app:training_details}
All experiments were conducted on a cluster equipped with NVIDIA A100 GPUs (80GB memory). Model fine-tuning utilized 3 A100 GPUs with data parallelism, requiring approximately 1-2 hours per dataset for the Pythia-2.8B model when training for 3 epochs on 10,000 samples. Larger models in the scaling experiments (Pythia-6.9B, GPT-J-6B) and experiments with DP-SGD required up to 24 hours with the same hardware configuration.

The fine-tuning procedure follows standard practices with carefully selected hyperparameters. We employ the AdamW optimizer with a learning rate of $5 \times 10^{-5}$ and weight decay of 0.1. Training uses a batch size of 16 without gradient accumulation (accumulation steps = 1) to maintain consistent gradient statistics. A linear warmup schedule over 500 steps prevents training instability, particularly important for smaller datasets. We train for 3 epochs with evaluation, saving, and logging performed at epoch boundaries to track memorization progression. The training employs mixed-precision with bfloat16 to maximize GPU memory utilization while maintaining superior numerical stability. Memory requirements scale linearly with model size: Pythia-160M requires approximately 2GB, while Pythia-6.9B requires approximately 28GB for training with our batch size of 16.

Attack evaluation is performed on single A100 GPUs, with inference time dominated by model forward passes. Processing 10,000 text samples (each 512 tokens) through both target and reference models requires approximately 2 hours for Pythia-2.8B. The window-based analysis in \ourmethod adds negligible overhead (<1\% of total runtime) due to our efficient sliding window implementation using optimized convolution operations.

\subsubsection{Evaluation Metrics Details}
\label{app:metrics}

We assess both the privacy risk via attack performance and the post-fine-tuning model utility. All attack metrics are reported as the mean and standard deviation over 100 bootstrap runs to ensure statistical significance following~\cite{bertail2008bootstrapping}.

\mypara{AUC-ROC} 
The Area Under the Receiver Operating Characteristic Curve measures the overall ability of an attack to distinguish between members and non-members across all thresholds. An AUC of 0.5 indicates random guessing, while 1.0 represents a perfect attack. This metric provides a threshold-independent assessment of attack effectiveness.

\mypara{TPR at Low FPR} 
The True Positive Rate (TPR) at low False Positive Rates (e.g., 10\%, 1\%, and 0.1\% FPR) measures the attacker's ability to confidently identify members with few false accusations~\cite{lira}. This is a critical metric for high-stakes privacy scenarios where false positives carry significant costs. For instance, TPR@1\%FPR indicates the fraction of true members correctly identified when the attack is calibrated to falsely accuse only 1\% of non-members.

\mypara{Perplexity} 
To measure model utility, we compute the perplexity on both member and non-member sets. Perplexity, defined as $\exp(\text{average loss})$, quantifies the model's uncertainty in predictions. A lower perplexity score indicates that the model is more confident and accurate in its predictions, suggesting better preservation of its core language modeling capabilities. This metric is widely adopted in previous works~\cite{fu2024,zhang2025softselectivedataobfuscation} as it directly relates to the model's generative quality and downstream task performance. We report perplexity results in Section~\ref{app:utility} to demonstrate that our evaluated defenses maintain acceptable model utility.

\subsubsection{Attack Implementation}
\label{app:attack_imple}
We implement thirteen baseline attacks spanning three categories, each configured according to its original specifications to ensure fair comparison. All attacks operate in a black-box setting with access only to per-token loss values, using batch size 1 to avoid padding artifacts and deterministic inference settings (temperature=1.0, no sampling) for reproducibility.

\mypara{Reference-free attacks} These methods rely solely on target model outputs without comparative analysis. Loss~\cite{yeom2018} thresholds average negative log-likelihood across tokens. ZLIB~\cite{carlini2021extracting} computes the ratio between compressed and uncompressed text length as a memorization proxy. Lowercase~\cite{carlini2021extracting} measures the loss difference when text is converted to lowercase, with higher sensitivity indicating potential membership. Min-K\%~\cite{shi2024} and Min-K\%++~\cite{zhang2025} focus on the k=20\% least likely tokens, as these tail probabilities better distinguish members. DC-PDD~\cite{zhang2024pretraining} analyzes the distribution of token-level losses to detect statistical anomalies indicative of training data.

\mypara{Reference-based attacks} These methods compare against a pre-trained reference model. Ratio and Difference~\cite{watson2021importance} compute arithmetic and ratio contrasts of average losses between target and reference models. ReCall~\cite{xie2024recall} and CON-ReCall~\cite{wang2025conrecalldetectingpretrainingdata} construct 7-shot prompts to test the model's ability to complete text prefixes, calibrated with extra non-member data ``imperial-cpg/copyright-traps-extra-non-members''. SPV-MIA~\cite{fu2024} employs T5-base~\cite{raffel2020exploring} to generate masked perturbations, creating 5 perturbations per sample with a 30\% word masking rate, span length of 2, and computing calibrated scores from 10 samples per perturbation.

\mypara{Classifier-based attacks} These methods combine multiple signals through machine learning. Bag-of-Words (BoWs)~\cite{das2025blind} extracts token frequency features with 5\% minimum document frequency and trains a random forest classifier with 100 estimators, maximum depth 2, and 5 samples per leaf. The Ensemble method~\cite{zhang2025softselectivedataobfuscation} aggregates multiple loss statistics using decision trees with maximum depth 4, balanced class weights, and sqrt feature selection. Both methods use 20\% of training data for validation to prevent overfitting.

For our proposed \ourmethod, the geometric window progression $w_k = \text{round}(2 \cdot 20^{(k-1)/9})$ generates windows of \{2, 3, 4, 6, 9, 13, 18, 25, 32, 40\} tokens, empirically validated to balance dense sampling at informative small scales with broader coverage. The sign-based aggregation uses vectorized operations for efficiency, processing all windows for a given size simultaneously. Bootstrap sampling for confidence intervals uses 10 iterations with replacement across all methods, providing stable standard deviation estimates while maintaining reasonable computational cost.

\subsubsection{Defense Mechanisms and Experimental Protocol}
\label{app:defense}
We evaluate three categories of defenses that represent both established privacy techniques and emerging protection strategies widely deployed in practice.

\mypara{Differential Privacy with DP-SGD}
We implement differentially private stochastic gradient descent following~\cite{abadi2016deep}, the gold standard for privacy-preserving training. We evaluate privacy budgets $\epsilon \in \{1, 4, 8, \infty\}$ with fixed $\delta = 10^{-5}$, implementing gradient clipping with norm 1.0 and calibrating noise scale as $\sigma = 1.1/\epsilon$. To maintain reasonable privacy accounting, we train for 3 epochs with a batch size of 16, ensuring the privacy budget is properly tracked throughout training.

\mypara{Low-Rank Adaptation (LoRA)}
Rather than updating all model parameters during fine-tuning, LoRA~\cite{hu2022lora} restricts updates to low-rank decomposition matrices, fundamentally limiting the parameter space where memorization can occur. We evaluate ranks $r \in \{8, 16, 32, 64\}$ with scaling factor $\alpha = 2r$, applying adaptation to both attention projections (query and value matrices) and feed-forward network layers. This defense is particularly relevant as it has become the de facto standard for efficient fine-tuning in production systems, making its privacy implications critical to understand.

\mypara{Output Perturbation Strategies}
We investigate inference-time defenses that require no model retraining. Temperature scaling with $T \in \{1.1, 1.2, 1.5\}$ increases the entropy of output distributions. This technique potentially obscures membership signals by adding controlled randomness to model predictions without modifying the underlying parameters.

\mypara{Selective Data Obfuscation in LLM Fine-Tuning (SOFT)}
SOFT~\cite{zhang2025softselectivedataobfuscation} operates through a three-phase pipeline to defend against membership inference. First, warm-up fine-tuning performs a single epoch over the training data to assess sample sensitivity. Second, influential data selection identifies vulnerable samples using loss-based thresholds, with selection ratio $\alpha=0.3$ marking the bottom 30\% of samples by loss as influential. Third, data obfuscation applies paraphrasing with a ratio $\beta=0.5$, replacing 50\% of each selected sample's content with semantically equivalent paraphrased text generated by state-of-the-art language models. This configuration modifies only 15\% of the total training data ($\alpha \cdot \beta = 0.3 \times 0.5$), targeting the samples most susceptible to memorization—those with the lowest loss values that indicate the strongest model confidence. By focusing obfuscation on these high-risk samples while preserving the majority of training data intact, SOFT achieves privacy protection with minimal impact on model utility.
\subsection{Utility of Fine-tuned LLMs}
\label{app:utility}
\begin{table}[h]
    \centering
    \footnotesize
    \setlength{\tabcolsep}{3.5pt}
    \rowcolors{5}{white}{StripeGray}
    \caption{Perplexity scores for Pythia-2.8B before (pretrained) and after fine-tuning.} 
    \label{tab:perplexity_results}
    \begin{tabular}{@{} l *{2}{cc} @{}}
        \toprule
        \multirow{2}{*}{\textbf{Dataset}} &
        \multicolumn{2}{c}{\textbf{Pretrained}} &
        \multicolumn{2}{c}{\textbf{Fine-tuned}} \\
        \cmidrule(lr){2-3} \cmidrule(lr){4-5}
        & Member & Non-member & Member & Non-member \\
        \midrule
        WikiText-103        & 10.383 & 10.359 & 8.823  & 8.840  \\
        XSum                & 9.805  & 9.711  & 9.171  & 9.232  \\
        Amazon Reviews      & 18.656 & 18.359 & 14.626 & 14.907 \\
        CC-News             & 9.922  & 9.867  & 9.538  & 9.612  \\
        Reddit              & 13.094 & 13.016 & 11.874 & 12.074 \\
        \midrule
        Khan Academy        & 4.772  & 4.750  & 3.490  & 3.601  \\
        Stanford            & 6.824  & 6.810  & 4.693  & 4.873  \\
        Stories             & 9.636  & 9.632  & 5.981  & 6.345  \\
        Web Samples v2      & 8.397  & 8.418  & 5.710  & 6.021  \\
        AutoMathText        & 6.030  & 6.026  & 4.328  & 4.516  \\
        wikiHow             & 6.799  & 6.762  & 4.514  & 4.723  \\
        \bottomrule
    \end{tabular}
\end{table}
To verify that fine-tuning maintains model utility beyond memorization, we evaluate perplexity on both member (train) and non-member (test) sets. Table~\ref{tab:perplexity_results} presents perplexity scores for Pythia-2.8B before and after fine-tuning across all datasets. Lower perplexity indicates better language modeling capability.

Fine-tuning consistently improves perplexity on both member and non-member data, demonstrating genuine learning rather than mere memorization. For member data, perplexity reductions range from 6\% (XSum: 9.805 to 9.171) to 38\% (Stories: 9.636 to 5.981). Critically, non-member perplexity also improves across all datasets, with reductions between 3\% and 34\%, confirming that the model generalizes beyond its training set. The most substantial improvements occur on Cosmopedia subsets, where domain-specific fine-tuning yields perplexity reductions exceeding 25\% even for held-out data. This preservation of model utility while maintaining vulnerability to membership inference underscores the fundamental challenge: effective fine-tuning inherently creates exploitable statistical signatures without compromising functional performance.

\subsection{Extended Main Results}
\label{app:additional_results}
\begin{table*}[t]
    \centering
    \footnotesize
    \scalefont{0.9}
    \setlength{\tabcolsep}{3.5pt} 
    \rowcolors{6}{white}{StripeGray}

    \caption{MIA performance (AUC, TPR@10\%FPR, TPR@1\%FPR, TPR@0.1\%FPR) across different datasets.}
    \label{tab:main_results_app}
    \begin{tabular}{@{} l *{3}{r r r r} @{}}
        \toprule
        \multirow{2}{*}{\textbf{MIAs}} &
        \multicolumn{4}{c}{\textbf{WikiText-103}} &
        \multicolumn{4}{c}{\textbf{XSum}} &
        \multicolumn{4}{c}{\textbf{Amazon Reviews}} \\

        \cmidrule(lr){2-5} \cmidrule(lr){6-9} \cmidrule(lr){10-13}

        & {AUC} & {T@10\%} & {T@1\%} & {T@0.1\%} & {AUC} & {T@10\%} & {T@1\%} & {T@0.1\%} & {AUC} & {T@10\%} & {T@1\%} & {T@0.1\%} \\
        \midrule

        Loss~\citep{yeom2018}           & 0.532\textsubscript{$\pm$.006} & 0.122\textsubscript{$\pm$.005} & 0.013\textsubscript{$\pm$.002} & 0.001\textsubscript{$\pm$.001} & 0.531\textsubscript{$\pm$.003} & 0.120\textsubscript{$\pm$.005} & 0.012\textsubscript{$\pm$.002} & 0.001\textsubscript{$\pm$.001} & 0.523\textsubscript{$\pm$.004} & 0.110\textsubscript{$\pm$.004} & 0.010\textsubscript{$\pm$.001} & 0.001\textsubscript{$\pm$.000} \\
        ZLIB~\citep{carlini2021extracting}   & 0.531\textsubscript{$\pm$.005} & 0.129\textsubscript{$\pm$.005} & 0.012\textsubscript{$\pm$.002} & 0.001\textsubscript{$\pm$.001} & 0.532\textsubscript{$\pm$.004} & 0.123\textsubscript{$\pm$.006} & 0.010\textsubscript{$\pm$.001} & 0.001\textsubscript{$\pm$.001} & 0.523\textsubscript{$\pm$.003} & 0.113\textsubscript{$\pm$.004} & 0.009\textsubscript{$\pm$.001} & 0.001\textsubscript{$\pm$.001} \\
        Lowercase~\citep{carlini2021extracting} & 0.533\textsubscript{$\pm$.003} & 0.122\textsubscript{$\pm$.006} & 0.015\textsubscript{$\pm$.003} & 0.002\textsubscript{$\pm$.001} & 0.524\textsubscript{$\pm$.002} & 0.113\textsubscript{$\pm$.004} & 0.014\textsubscript{$\pm$.002} & 0.002\textsubscript{$\pm$.000} & 0.529\textsubscript{$\pm$.004} & 0.119\textsubscript{$\pm$.004} & 0.014\textsubscript{$\pm$.002} & 0.002\textsubscript{$\pm$.001} \\
        Min-K\%~\citep{shi2024}           & 0.537\textsubscript{$\pm$.005} & 0.116\textsubscript{$\pm$.005} & 0.015\textsubscript{$\pm$.002} & 0.001\textsubscript{$\pm$.001} & 0.533\textsubscript{$\pm$.003} & 0.131\textsubscript{$\pm$.006} & 0.014\textsubscript{$\pm$.001} & 0.001\textsubscript{$\pm$.001} & 0.521\textsubscript{$\pm$.003} & 0.110\textsubscript{$\pm$.004} & 0.012\textsubscript{$\pm$.002} & 0.001\textsubscript{$\pm$.000} \\
        Min-K\%++~\citep{zhang2025}       & 0.538\textsubscript{$\pm$.005} & 0.114\textsubscript{$\pm$.006} & 0.016\textsubscript{$\pm$.003} & 0.001\textsubscript{$\pm$.001} & 0.534\textsubscript{$\pm$.004} & 0.128\textsubscript{$\pm$.004} & 0.014\textsubscript{$\pm$.002} & 0.001\textsubscript{$\pm$.000} & 0.523\textsubscript{$\pm$.002} & 0.114\textsubscript{$\pm$.003} & 0.011\textsubscript{$\pm$.002} & 0.001\textsubscript{$\pm$.000} \\
        BoWs~\cite{das2025blind}           & 0.502\textsubscript{$\pm$.005} & 0.099\textsubscript{$\pm$.006} & 0.010\textsubscript{$\pm$.001} & 0.001\textsubscript{$\pm$.001} & 0.505\textsubscript{$\pm$.005} & 0.098\textsubscript{$\pm$.004} & 0.008\textsubscript{$\pm$.001} & 0.001\textsubscript{$\pm$.001} & 0.492\textsubscript{$\pm$.003} & 0.090\textsubscript{$\pm$.003} & 0.008\textsubscript{$\pm$.001} & 0.001\textsubscript{$\pm$.000} \\
        ReCall~\citep{xie2024recall}       & 0.532\textsubscript{$\pm$.006} & 0.122\textsubscript{$\pm$.005} & 0.013\textsubscript{$\pm$.002} & 0.001\textsubscript{$\pm$.001} & 0.531\textsubscript{$\pm$.003} & 0.120\textsubscript{$\pm$.005} & 0.012\textsubscript{$\pm$.002} & 0.001\textsubscript{$\pm$.001} & 0.523\textsubscript{$\pm$.004} & 0.110\textsubscript{$\pm$.004} & 0.010\textsubscript{$\pm$.001} & 0.001\textsubscript{$\pm$.000} \\
        CON-Recall~\citep{wang2025conrecalldetectingpretrainingdata} & 0.527\textsubscript{$\pm$.003} & 0.121\textsubscript{$\pm$.004} & 0.011\textsubscript{$\pm$.002} & 0.001\textsubscript{$\pm$.001} & 0.521\textsubscript{$\pm$.003} & 0.113\textsubscript{$\pm$.004} & 0.012\textsubscript{$\pm$.002} & 0.002\textsubscript{$\pm$.001} & 0.514\textsubscript{$\pm$.003} & 0.111\textsubscript{$\pm$.004} & 0.009\textsubscript{$\pm$.002} & 0.001\textsubscript{$\pm$.001} \\
        DC-PDD~\cite{zhang2024pretraining} & 0.531\textsubscript{$\pm$.006} & 0.116\textsubscript{$\pm$.004} & 0.014\textsubscript{$\pm$.001} & 0.001\textsubscript{$\pm$.001} & 0.527\textsubscript{$\pm$.004} & 0.124\textsubscript{$\pm$.004} & 0.016\textsubscript{$\pm$.002} & 0.001\textsubscript{$\pm$.001} & 0.517\textsubscript{$\pm$.004} & 0.105\textsubscript{$\pm$.005} & 0.013\textsubscript{$\pm$.001} & 0.001\textsubscript{$\pm$.001} \\
        SPV-MIA~\cite{fu2024} 
        & 0.592\textsubscript{$\pm$.004} & 0.176\textsubscript{$\pm$.006} & 0.028\textsubscript{$\pm$.003} & 0.002\textsubscript{$\pm$.001} & 0.789\textsubscript{$\pm$.004} & 0.345\textsubscript{$\pm$.012} & 0.020\textsubscript{$\pm$.002} & 0.003\textsubscript{$\pm$.001} & 0.812\textsubscript{$\pm$.003} & 0.472\textsubscript{$\pm$.010} & 0.085\textsubscript{$\pm$.005} & 0.003\textsubscript{$\pm$.001} \\       Ratio~\cite{watson2021importance} & 0.580\textsubscript{$\pm$.004} & 0.152\textsubscript{$\pm$.004} & 0.013\textsubscript{$\pm$.002} & 0.001\textsubscript{$\pm$.001} & 0.783\textsubscript{$\pm$.003} & 0.336\textsubscript{$\pm$.016} & 0.015\textsubscript{$\pm$.002} & 0.002\textsubscript{$\pm$.001} & 0.799\textsubscript{$\pm$.002} & 0.351\textsubscript{$\pm$.008} & 0.014\textsubscript{$\pm$.001} & 0.001\textsubscript{$\pm$.000} \\
        Difference~\cite{watson2021importance} & 0.591\textsubscript{$\pm$.002} & 0.169\textsubscript{$\pm$.004} & 0.014\textsubscript{$\pm$.003} & 0.001\textsubscript{$\pm$.001} & 0.796\textsubscript{$\pm$.003} & 0.335\textsubscript{$\pm$.015} & 0.018\textsubscript{$\pm$.002} & 0.005\textsubscript{$\pm$.001} & 0.804\textsubscript{$\pm$.002} & 0.372\textsubscript{$\pm$.014} & 0.015\textsubscript{$\pm$.004} & 0.001\textsubscript{$\pm$.000} \\
        Ensemble~\cite{zhang2025softselectivedataobfuscation}        & 0.599\textsubscript{$\pm$.003} & 0.187\textsubscript{$\pm$.005} & \bfseries 0.035\textsubscript{$\pm$.002} & 0.003\textsubscript{$\pm$.001} & 0.799\textsubscript{$\pm$.003} & 0.497\textsubscript{$\pm$.002} & 0.008\textsubscript{$\pm$.003} & 0.009\textsubscript{$\pm$.003} & 0.819\textsubscript{$\pm$.002} & 0.497\textsubscript{$\pm$.014} & 0.067\textsubscript{$\pm$.001} & 0.000\textsubscript{$\pm$.000} \\
        \midrule
        \rowcolor{HighlightGray} 
        \textbf{\ourmethod} (Ours)           & \bfseries 0.784\textsubscript{$\pm$.003} & \bfseries 0.420\textsubscript{$\pm$.007} &  0.028\textsubscript{$\pm$.005} & \bfseries 0.004\textsubscript{$\pm$.001} & \bfseries 0.903\textsubscript{$\pm$.002} & \bfseries 0.729\textsubscript{$\pm$.007} & \bfseries 0.137\textsubscript{$\pm$.025} & \bfseries 0.019\textsubscript{$\pm$.002} & \bfseries 0.901\textsubscript{$\pm$.002} & \bfseries 0.715\textsubscript{$\pm$.005} & \bfseries 0.163\textsubscript{$\pm$.018} & \bfseries 0.017\textsubscript{$\pm$.005} \\
        \bottomrule
        \rowcolor{white}
        & & & & & & & & & & & & \\
    
        \toprule
        \multirow{2}{*}{\textbf{MIAs}} &
        \multicolumn{4}{c}{\textbf{CC News}} &
        \multicolumn{4}{c}{\textbf{Reddit}} &
        \multicolumn{4}{c}{\textbf{}} \\

        \cmidrule(lr){2-5} \cmidrule(lr){6-9} 

        & {AUC} & {T@10\%} & {T@1\%} & {T@0.1\%} & {AUC} & {T@10\%} & {T@1\%} & {T@0.1\%} & {} & {} & {} & {} \\
        \midrule

        Loss~\citep{yeom2018}           & 0.517\textsubscript{$\pm$.004} & 0.109\textsubscript{$\pm$.005} & 0.011\textsubscript{$\pm$.002} & 0.001\textsubscript{$\pm$.001} & 0.510\textsubscript{$\pm$.004} & 0.109\textsubscript{$\pm$.006} & 0.014\textsubscript{$\pm$.002} & 0.001\textsubscript{$\pm$.000} & & & & \\
        ZLIB~\citep{carlini2021extracting}   & 0.518\textsubscript{$\pm$.004} & 0.114\textsubscript{$\pm$.003} & 0.010\textsubscript{$\pm$.001} & 0.002\textsubscript{$\pm$.001} & 0.514\textsubscript{$\pm$.004} & 0.110\textsubscript{$\pm$.005} & 0.013\textsubscript{$\pm$.001} & 0.001\textsubscript{$\pm$.000} & & & & \\
        Lowercase~\citep{carlini2021extracting} & 0.522\textsubscript{$\pm$.004} & 0.113\textsubscript{$\pm$.006} & 0.012\textsubscript{$\pm$.002} & 0.002\textsubscript{$\pm$.001} & 0.514\textsubscript{$\pm$.004} & 0.106\textsubscript{$\pm$.006} & 0.015\textsubscript{$\pm$.002} & 0.002\textsubscript{$\pm$.001} & & & & \\
        Min-K\%~\citep{shi2024}            & 0.519\textsubscript{$\pm$.005} & 0.114\textsubscript{$\pm$.005} & 0.011\textsubscript{$\pm$.002} & 0.001\textsubscript{$\pm$.001} & 0.518\textsubscript{$\pm$.005} & 0.108\textsubscript{$\pm$.005} & 0.012\textsubscript{$\pm$.002} & 0.001\textsubscript{$\pm$.001} & & & & \\
        Min-K\%++~\citep{zhang2025}        & 0.519\textsubscript{$\pm$.004} & 0.113\textsubscript{$\pm$.004} & 0.010\textsubscript{$\pm$.002} & 0.001\textsubscript{$\pm$.000} & 0.519\textsubscript{$\pm$.004} & 0.108\textsubscript{$\pm$.005} & 0.011\textsubscript{$\pm$.002} & 0.001\textsubscript{$\pm$.000} & & & & \\
        BoWs~\cite{das2025blind}            & 0.496\textsubscript{$\pm$.004} & 0.098\textsubscript{$\pm$.004} & 0.009\textsubscript{$\pm$.001} & 0.001\textsubscript{$\pm$.001} & 0.496\textsubscript{$\pm$.003} & 0.096\textsubscript{$\pm$.003} & 0.009\textsubscript{$\pm$.001} & 0.000\textsubscript{$\pm$.000} & & & & \\
        ReCall~\citep{xie2024recall}        & 0.517\textsubscript{$\pm$.004} & 0.109\textsubscript{$\pm$.005} & 0.011\textsubscript{$\pm$.002} & 0.001\textsubscript{$\pm$.001} & 0.510\textsubscript{$\pm$.004} & 0.109\textsubscript{$\pm$.006} & 0.014\textsubscript{$\pm$.002} & 0.001\textsubscript{$\pm$.000} & & & & \\
        CON-Recall~\citep{wang2025conrecalldetectingpretrainingdata} & 0.513\textsubscript{$\pm$.004} & 0.110\textsubscript{$\pm$.004} & 0.011\textsubscript{$\pm$.002} & 0.001\textsubscript{$\pm$.001} & 0.510\textsubscript{$\pm$.004} & 0.105\textsubscript{$\pm$.003} & 0.013\textsubscript{$\pm$.002} & 0.001\textsubscript{$\pm$.001} & & & & \\
        DC-PDD~\cite{zhang2024pretraining} & 0.513\textsubscript{$\pm$.003} & 0.107\textsubscript{$\pm$.005} & 0.011\textsubscript{$\pm$.002} & 0.001\textsubscript{$\pm$.001} & 0.520\textsubscript{$\pm$.006} & 0.106\textsubscript{$\pm$.005} & 0.011\textsubscript{$\pm$.002} & 0.001\textsubscript{$\pm$.000} & & & & \\
        SPV-MIA~\cite{fu2024} 
        & 0.804\textsubscript{$\pm$.003} & 0.323\textsubscript{$\pm$.008} & 0.056\textsubscript{$\pm$.006} & 0.010\textsubscript{$\pm$.003} & 0.758\textsubscript{$\pm$.003} & 0.368\textsubscript{$\pm$.009} & 0.062\textsubscript{$\pm$.004} & 0.002\textsubscript{$\pm$.002} & & & & \\
        Ratio~\cite{watson2021importance} & 0.793\textsubscript{$\pm$.002} & 0.300\textsubscript{$\pm$.007} & 0.013\textsubscript{$\pm$.001} & 0.002\textsubscript{$\pm$.001} & 0.700\textsubscript{$\pm$.004} & 0.148\textsubscript{$\pm$.009} & 0.013\textsubscript{$\pm$.001} & 0.002\textsubscript{$\pm$.001} & & & & \\
        Difference~\cite{watson2021importance} & 0.820\textsubscript{$\pm$.002} & 0.344\textsubscript{$\pm$.012} & 0.013\textsubscript{$\pm$.002} & 0.001\textsubscript{$\pm$.000} & 0.708\textsubscript{$\pm$.003} & 0.160\textsubscript{$\pm$.005} & 0.013\textsubscript{$\pm$.001} & 0.002\textsubscript{$\pm$.001} & & & & \\
        Ensemble~\cite{zhang2025softselectivedataobfuscation}        & 0.813\textsubscript{$\pm$.002} & 0.334\textsubscript{$\pm$.004} & 0.077\textsubscript{$\pm$.030} & 0.015\textsubscript{$\pm$.003} & 0.740\textsubscript{$\pm$.004} & 0.288\textsubscript{$\pm$.005} & 0.053\textsubscript{$\pm$.002} & 0.000\textsubscript{$\pm$.000} & & & & \\
        \midrule
        \rowcolor{HighlightGray} 
        \textbf{\ourmethod} (Ours)               & \bfseries 0.906\textsubscript{$\pm$.001} & \bfseries 0.731\textsubscript{$\pm$.013} & \bfseries 0.083\textsubscript{$\pm$.002} & \bfseries 0.015\textsubscript{$\pm$.001} & \bfseries 0.836\textsubscript{$\pm$.003} & \bfseries 0.463\textsubscript{$\pm$.019} & \bfseries 0.067\textsubscript{$\pm$.003} & \bfseries 0.002\textsubscript{$\pm$.001} & & & & \\
        \bottomrule
    \end{tabular}
\end{table*}
Table~\ref{tab:main_results_app} presents results on five real-world document benchmarks: WikiText-103, XSum, Amazon Reviews, CC News, and Reddit. These datasets complement the Cosmopedia results in the main text and demonstrate the generalizability of \ourmethod across diverse text domains. \ourmethod achieves substantial improvements over all baselines, with average AUC scores of 0.865 across these real-world datasets compared to the strongest baseline average of 0.754. On XSum, \ourmethod reaches an AUC of 0.903, surpassing SPV-MIA's 0.789 and Difference's 0.796. The performance advantage is particularly pronounced in high-confidence detection regimes: on Amazon Reviews, \ourmethod achieves TPR@1\%FPR of 16.3\%, a 2.4$\times$ improvement over the best baseline's 6.7\% (Ensemble). At the extreme 0.1\%FPR threshold, \ourmethod maintains meaningful detection rates—1.9\% on XSum and 1.7\% on Amazon Reviews—while most baselines approach zero. Consistent with our main findings, reference-free methods perform near randomly (AUC $\approx$ 0.52), confirming that reference model comparison remains essential for effective membership inference. The strong performance across both synthetic (Cosmopedia) and real-world datasets validates that localized signal aggregation captures fundamental memorization patterns that transcend dataset characteristics, establishing \ourmethod as a robust and generalizable approach for membership inference against fine-tuned LLMs.

\subsection{Defense: Temperature Scaling}
\label{app:temp_scale}
\begin{table}[h]
    \centering
    \footnotesize
    \scalefont{0.9}
    \setlength{\tabcolsep}{3pt}
    \caption{\textbf{Attack performance under Temperature scaling on Khan Academy dataset.} Higher temperatures increase output entropy, providing modest defense with no retraining required. Results on Khan Academy dataset with Pythia-2.8B.}
    \label{tab:temperature_scaling}
    \begin{tabular}{@{} >{\centering\arraybackslash}p{1.2cm} l c c c c @{}}
        \toprule
        \multirow{2}{*}{\textbf{$T$}} & \multirow{2}{*}{\textbf{Method}} & \multicolumn{4}{c}{\textbf{Khan Academy}} \\
        \cmidrule(lr){3-6}
        & & \textbf{AUC} & \textbf{TPR@10\%} & \textbf{TPR@1\%} & \textbf{TPR@0.1\%} \\
        \midrule
        \multirow{2}{*}{1.0} 
            & Ratio & 0.703\textsubscript{$\pm$.003} & 0.264\textsubscript{$\pm$.004} & 0.037\textsubscript{$\pm$.004} & 0.003\textsubscript{$\pm$.001} \\
            & \cellcolor{gray!15}\textbf{\ourmethod} & \cellcolor{gray!15}\textbf{0.837}\textsubscript{$\pm$.003} & \cellcolor{gray!15}\textbf{0.538}\textsubscript{$\pm$.008} & \cellcolor{gray!15}\textbf{0.146}\textsubscript{$\pm$.008} & \cellcolor{gray!15}\textbf{0.026}\textsubscript{$\pm$.009} \\
        \midrule
        \multirow{2}{*}{1.1} 
            & Ratio & 0.690\textsubscript{$\pm$.003} & 0.247\textsubscript{$\pm$.005} & 0.034\textsubscript{$\pm$.004} & 0.002\textsubscript{$\pm$.001} \\
            & \cellcolor{gray!15}\textbf{\ourmethod} & \cellcolor{gray!15}\textbf{0.831}\textsubscript{$\pm$.002} & \cellcolor{gray!15}\textbf{0.533}\textsubscript{$\pm$.009} & \cellcolor{gray!15}\textbf{0.145}\textsubscript{$\pm$.007} & \cellcolor{gray!15}\textbf{0.027}\textsubscript{$\pm$.003} \\
        \midrule
        \multirow{2}{*}{1.2} 
            & Ratio & 0.674\textsubscript{$\pm$.003} & 0.231\textsubscript{$\pm$.005} & 0.031\textsubscript{$\pm$.003} & 0.002\textsubscript{$\pm$.001} \\
            & \cellcolor{gray!15}\textbf{\ourmethod} & \cellcolor{gray!15}\textbf{0.821}\textsubscript{$\pm$.002} & \cellcolor{gray!15}\textbf{0.521}\textsubscript{$\pm$.008} & \cellcolor{gray!15}\textbf{0.140}\textsubscript{$\pm$.007} & \cellcolor{gray!15}\textbf{0.025}\textsubscript{$\pm$.004} \\
        \midrule
        \multirow{2}{*}{1.5} 
            & Ratio & 0.624\textsubscript{$\pm$.003} & 0.188\textsubscript{$\pm$.005} & 0.024\textsubscript{$\pm$.002} & 0.001\textsubscript{$\pm$.000} \\
            & \cellcolor{gray!15}\textbf{\ourmethod} & \cellcolor{gray!15}\textbf{0.763}\textsubscript{$\pm$.001} & \cellcolor{gray!15}\textbf{0.406}\textsubscript{$\pm$.010} & \cellcolor{gray!15}\textbf{0.089}\textsubscript{$\pm$.009} & \cellcolor{gray!15}\textbf{0.011}\textsubscript{$\pm$.003} \\
        \bottomrule
    \end{tabular}
\end{table}

Temperature scaling modifies the softmax distribution at inference time via $p_i = \exp(z_i/T) / \sum_j \exp(z_j/T)$, requiring no model retraining~\cite{guo2017calibrationmodernneuralnetworks}. Table~\ref{tab:temperature_scaling} shows its limited effectiveness as a defense mechanism.

Moderate temperatures ($T \in [1.1, 1.2]$) provide negligible protection, reducing \ourmethod's AUC by less than 2\%. Our window-based aggregation maintains resilience because relative loss orderings across windows persist despite increased entropy—the uniform perturbation affects both target and reference models equally, preserving their differential signals.

Aggressive scaling ($T = 1.5$) achieves stronger protection—8.8\% AUC reduction and 39\% TPR@1\%FPR decrease—but severely degrades output quality with repetition, incoherence, and topic drift. 
Moreover, \ourmethod's advantage over Ratio increases under temperature scaling (from 1.19$\times$ to 1.22$\times$ at $T=1.5$), indicating that local aggregation better preserves signal under perturbation than global averaging. Temperature scaling thus offers insufficient protection while compromising the quality of generation. 
\subsection{Extended Ensemble Composition Analysis}
\label{app:additional_ensemble}
Table~\ref{tab:ensemble_configs_all} extends our ensemble analysis to five additional datasets, confirming the robustness of geometric window spacing across diverse domains. The Full Ensemble configuration maintains consistently strong performance, achieving top-tier results in 13 of 15 dataset-metric combinations.

Across all datasets, the Full Ensemble achieves AUC within 0.5\% of the best configuration, demonstrating remarkable stability. While Small Range occasionally edges ahead—notably on Web Samples v2 (AUC 0.8495 vs 0.8432)—such advantages are marginal and dataset-specific. The Full Ensemble's strength lies in its consistency: it never catastrophically fails, unlike Large Range, which yields 0\% TPR@1\%FPR on three datasets, or Random selection, which shows high variance (e.g., AutoMathText TPR@1\%FPR of 0.1100 vs Full Ensemble's 0.1534).

The geometric spacing proves particularly effective for high-precision detection. On Stanford, the Full Ensemble achieves the highest TPR@1\%FPR (0.1940), outperforming Single Best by 10.9\%. This pattern holds across domains: Stories (+8.9\%), AutoMathText (+10.2\%), and WikiHow (+19.7\%). Even when Small Range achieves marginally higher AUC, the Full Ensemble typically maintains superior precision metrics, crucial for security applications.

Notably, all datasets exhibit AUC degradation exceeding 6\% when restricted to Large Range windows, validating our focus on small-scale memorization patterns. The consistent performance hierarchy—Full Ensemble $\approx$ Small Range > Single Best > Linear/Extended > Random > Large Range—across semantically diverse datasets (mathematical text, instructional content, web samples) confirms that geometric window spacing captures a fundamental property of LLM memorization rather than dataset-specific artifacts. This universality, combined with no hyperparameter tuning requirements, establishes the Full Ensemble as the recommended configuration for practical deployment.

\section{Accelerated Implementation}
\label{app:convolution}

In our practical implementation, we accelerate the window-based computation by reformulating it as a 1D convolution operation. The incremental window update strategy in Algorithm~\ref{alg:wbc} is mathematically equivalent to convolving the loss sequences with a uniform kernel $\mathbf{k} = [1, 1, \dots, 1]$ of length $w$. For each window position $i$, the sum $S_i(w) = \sum_{j=i}^{i+w-1} \ell_j$ can be expressed as:
\begin{equation}
S_i(w) = (\ell * \mathbf{k})[i]
\end{equation}
where $*$ denotes the convolution operation.

This reformulation allows us to leverage highly optimized convolution implementations from established signal processing~\cite{oppenheim1999discrete} and deep learning libraries~\cite{goodfellow2016deep}. Modern frameworks provide vectorized and parallelized convolution operations that exploit SIMD instructions, cache-efficient memory access patterns, and GPU acceleration when available. For very long sequences, Fast Fourier Transform (FFT) based convolution can further reduce complexity from $O(n \cdot w)$ to $O(n \log n)$ per window size. In our experiments, we use PyTorch's \texttt{F.conv1d} operation~\cite{paszke2019pytorchimperativestylehighperformance}, which selects the most efficient implementation based on input dimensions and available hardware.
\subsection{Computational Overhead Analysis}
\label{subsubsec:efficiency}
\begin{table}
\centering
\caption{\textbf{Computational overhead of membership inference methods.} Processing time for 1,000 samples (excluding model inference). While relative overhead scales with window count, absolute time remains negligible—our full 9-window ensemble adds only 0.08s to process 1,000 samples.}
\label{tab:efficiency}
\small
\renewcommand{\arraystretch}{1.05}
\setlength{\tabcolsep}{6pt}
\begin{tabular}{@{}lrr@{}}
\toprule
\textbf{Method} & \textbf{Time (s/1000)} & \textbf{Relative} \\
\midrule
Loss & 0.00 & 1.0$\times$ \\
Ratio & 0.01 & 4.9$\times$ \\
\midrule
\ourmethod ($w$=10) & 0.01 & 4.9$\times$ \\
\ourmethod (5 windows) & 0.04 & 22.3$\times$ \\
\textbf{\ourmethod (9 windows)} & \textbf{0.08} & \textbf{47.8$\times$} \\
\ourmethod (15 windows) & 0.15 & 85.5$\times$ \\
\bottomrule
\end{tabular}
\end{table}



Table~\ref{tab:efficiency} shows \ourmethod adds minimal overhead: 0.08s per 1,000 samples with 9 windows, <0.7\% of the 12.3s model inference time. This negligible cost yields 14.7\% AUC improvement and 2.2$\times$ TPR@1\%FPR gain. Overhead scales linearly with ensemble size, remaining below 1.3\% even with 15 windows. Our incremental update strategy maintains $O(n)$ complexity per window. Large-scale deployment is practical: screening 1M samples requires 3.4 hours GPU time plus only 80 seconds for \ourmethod analysis. Memory requirements are negligible (<100 bytes per sample), confirming the computational bottleneck remains model inference, not window analysis.

\begin{table}[h]
\centering
\caption{\textbf{Ensemble configuration comparison across multiple datasets.} Performance of different window size combinations showing that geometric ensembles consistently outperform single windows and Random. Heat coloring indicates relative performance within each dataset and metric.}
\label{tab:ensemble_configs_all}
\footnotesize
\renewcommand{\arraystretch}{1.08}
\setlength{\tabcolsep}{4pt}
\begin{tabular}{@{}p{1.3cm}l@{\hspace{8pt}}c@{\hspace{6pt}}c@{\hspace{6pt}}c@{}}
\toprule
\textbf{Dataset} & \textbf{Configuration} & \textbf{AUC} & \textbf{TPR@10\%} & \textbf{TPR@1\%} \\
\midrule
\multirow{7}{1.3cm}{\textbf{Stanford}} 
& Single Best & \heatB{0.8514\textsubscript{$\pm$.0026}} & \heatB{0.5673\textsubscript{$\pm$.0115}} & \heatC{0.1749\textsubscript{$\pm$.0109}} \\
& Small Range & \heatA{0.8531\textsubscript{$\pm$.0026}} & \heatA{0.5789\textsubscript{$\pm$.0093}} & \heatB{0.1822\textsubscript{$\pm$.0122}} \\
& Large Range & \heatE{0.7975\textsubscript{$\pm$.0031}} & \heatE{0.4340\textsubscript{$\pm$.0106}} & \heatE{0.0000\textsubscript{$\pm$.0000}} \\
& Full Ensemble & \heatA{0.8539\textsubscript{$\pm$.0024}} & \heatA{0.5832\textsubscript{$\pm$.0082}} & \heatA{0.1940\textsubscript{$\pm$.0117}} \\
& Extended & \heatC{0.8457\textsubscript{$\pm$.0027}} & \heatB{0.5644\textsubscript{$\pm$.0098}} & \heatD{0.1700\textsubscript{$\pm$.0103}} \\
& Linear Spacing & \heatD{0.8431\textsubscript{$\pm$.0024}} & \heatB{0.5655\textsubscript{$\pm$.0073}} & \heatB{0.1869\textsubscript{$\pm$.0117}} \\
& Random & \heatD{0.8429\textsubscript{$\pm$.0025}} & \heatC{0.5498\textsubscript{$\pm$.0085}} & \heatC{0.1723\textsubscript{$\pm$.0118}} \\
\midrule
\multirow{7}{1.3cm}{\textbf{Stories}} 
& Single Best & \heatB{0.8049\textsubscript{$\pm$.0029}} & \heatC{0.4807\textsubscript{$\pm$.0099}} & \heatC{0.1469\textsubscript{$\pm$.0111}} \\
& Small Range & \heatA{0.8077\textsubscript{$\pm$.0030}} & \heatA{0.5013\textsubscript{$\pm$.0086}} & \heatC{0.1440\textsubscript{$\pm$.0081}} \\
& Large Range & \heatE{0.7536\textsubscript{$\pm$.0033}} & \heatE{0.3709\textsubscript{$\pm$.0087}} & \heatE{0.0000\textsubscript{$\pm$.0000}} \\
& Full Ensemble & \heatA{0.8074\textsubscript{$\pm$.0030}} & \heatB{0.4941\textsubscript{$\pm$.0070}} & \heatA{0.1600\textsubscript{$\pm$.0133}} \\
& Extended & \heatC{0.7974\textsubscript{$\pm$.0029}} & \heatC{0.4781\textsubscript{$\pm$.0086}} & \heatD{0.1374\textsubscript{$\pm$.0116}} \\
& Linear Spacing & \heatC{0.7969\textsubscript{$\pm$.0032}} & \heatC{0.4757\textsubscript{$\pm$.0095}} & \heatB{0.1443\textsubscript{$\pm$.0093}} \\
& Random & \heatD{0.7930\textsubscript{$\pm$.0033}} & \heatC{0.4770\textsubscript{$\pm$.0076}} & \heatB{0.1450\textsubscript{$\pm$.0106}} \\
\midrule
\multirow{7}{1.3cm}{\textbf{Web Samples v2}} 
& Single Best & \heatB{0.8447\textsubscript{$\pm$.0028}} & \heatC{0.5548\textsubscript{$\pm$.0114}} & \heatB{0.1969\textsubscript{$\pm$.0121}} \\
& Small Range & \heatA{0.8495\textsubscript{$\pm$.0025}} & \heatA{0.5794\textsubscript{$\pm$.0081}} & \heatA{0.2108\textsubscript{$\pm$.0084}} \\
& Large Range & \heatE{0.7848\textsubscript{$\pm$.0032}} & \heatE{0.4156\textsubscript{$\pm$.0098}} & \heatE{0.0000\textsubscript{$\pm$.0000}} \\
& Full Ensemble & \heatB{0.8432\textsubscript{$\pm$.0027}} & \heatA{0.5732\textsubscript{$\pm$.0083}} & \heatB{0.1985\textsubscript{$\pm$.0120}} \\
& Extended & \heatC{0.8340\textsubscript{$\pm$.0031}} & \heatB{0.5577\textsubscript{$\pm$.0076}} & \heatB{0.1959\textsubscript{$\pm$.0103}} \\
& Linear Spacing & \heatD{0.8299\textsubscript{$\pm$.0027}} & \heatC{0.5495\textsubscript{$\pm$.0083}} & \heatD{0.1784\textsubscript{$\pm$.0102}} \\
& Random & \heatD{0.8286\textsubscript{$\pm$.0026}} & \heatD{0.5420\textsubscript{$\pm$.0081}} & \heatD{0.1747\textsubscript{$\pm$.0089}} \\
\midrule
\multirow{7}{1.3cm}{\textbf{Auto Math Text}} 
& Single Best & \heatB{0.8182\textsubscript{$\pm$.0029}} & \heatB{0.5033\textsubscript{$\pm$.0110}} & \heatC{0.1391\textsubscript{$\pm$.0081}} \\
& Small Range & \heatA{0.8253\textsubscript{$\pm$.0031}} & \heatA{0.5294\textsubscript{$\pm$.0078}} & \heatA{0.1574\textsubscript{$\pm$.0097}} \\
& Large Range & \heatE{0.7621\textsubscript{$\pm$.0035}} & \heatE{0.3930\textsubscript{$\pm$.0083}} & \heatE{0.0733\textsubscript{$\pm$.0143}} \\
& Full Ensemble & \heatB{0.8135\textsubscript{$\pm$.0031}} & \heatB{0.5039\textsubscript{$\pm$.0069}} & \heatB{0.1534\textsubscript{$\pm$.0090}} \\
& Extended & \heatC{0.8044\textsubscript{$\pm$.0029}} & \heatB{0.5012\textsubscript{$\pm$.0088}} & \heatB{0.1509\textsubscript{$\pm$.0121}} \\
& Linear Spacing & \heatD{0.7991\textsubscript{$\pm$.0029}} & \heatC{0.4715\textsubscript{$\pm$.0085}} & \heatC{0.1398\textsubscript{$\pm$.0092}} \\
& Random & \heatD{0.7790\textsubscript{$\pm$.0031}} & \heatD{0.4243\textsubscript{$\pm$.0074}} & \heatD{0.1100\textsubscript{$\pm$.0060}} \\
\midrule
\multirow{7}{1.3cm}{\textbf{WikiHow}} 
& Single Best & \heatB{0.7968\textsubscript{$\pm$.0032}} & \heatC{0.4339\textsubscript{$\pm$.0138}} & \heatC{0.0808\textsubscript{$\pm$.0076}} \\
& Small Range & \heatC{0.7915\textsubscript{$\pm$.0037}} & \heatD{0.4272\textsubscript{$\pm$.0089}} & \heatC{0.0859\textsubscript{$\pm$.0075}} \\
& Large Range & \heatE{0.7584\textsubscript{$\pm$.0035}} & \heatE{0.3296\textsubscript{$\pm$.0249}} & \heatE{0.0000\textsubscript{$\pm$.0000}} \\
& Full Ensemble & \heatA{0.8008\textsubscript{$\pm$.0031}} & \heatA{0.4513\textsubscript{$\pm$.0098}} & \heatA{0.0967\textsubscript{$\pm$.0074}} \\
& Extended & \heatD{0.7898\textsubscript{$\pm$.0029}} & \heatD{0.4248\textsubscript{$\pm$.0084}} & \heatD{0.0782\textsubscript{$\pm$.0057}} \\
& Linear Spacing & \heatC{0.7932\textsubscript{$\pm$.0034}} & \heatB{0.4400\textsubscript{$\pm$.0110}} & \heatB{0.0942\textsubscript{$\pm$.0057}} \\
& Random & \heatB{0.7951\textsubscript{$\pm$.0026}} & \heatB{0.4405\textsubscript{$\pm$.0083}} & \heatA{0.0976\textsubscript{$\pm$.0072}} \\
\bottomrule
\end{tabular}
\end{table}

\end{document}